# OBSUM: An object-based spatial unmixing model for spatiotemporal fusion of remote sensing images

Houcai Guo [a], Dingqi Ye [b], Lorenzo Bruzzone [a, *]

[a] Department of Information Engineering and Computer Science, University of Trento, Trento 38123, Italy

[b] School of Geosciences and Info-Physics, Central South University, Changsha 410083, China

[*] Corresponding author. E-mail: lorenzo.bruzzone@unitn.it

**Highlights:**

- OBSUM is proposed for the spatiotemporal fusion of remote sensing images.
- OBSUM considers object-level information of the land surface.
- OBSUM requires a minimum number of input images.
- OBSUM can retrieve strong temporal changes and obtain high fusion accuracy.
- OBSUM can support various remote sensing applications.

**Abstract:** Spatiotemporal fusion aims to improve both the spatial and temporal resolution of remote sensing images, thus facilitating time-series analysis at a fine spatial scale. However, there are several important issues that limit the application of current spatiotemporal fusion methods. First, most spatiotemporal fusion methods are based on pixel-level computation, which neglects the valuable object-level information of the land surface. Moreover, many existing methods cannot accurately retrieve strong temporal changes between the available high-resolution image at base date and the predicted one. This study proposes an Object-Based Spatial Unmixing Model (OBSUM), which incorporates object-based image analysis and spatial unmixing, to overcome the two abovementioned problems. OBSUM consists of one preprocessing step and three fusion steps, i.e., object-level unmixing, object-level residual compensation, and pixel-level residual compensation. OBSUM can be applied using only one fine image at the base date and one coarse image at the prediction date, without the need of a coarse image at the base date. The performance of OBSUM was compared with five representative spatiotemporal fusion methods. The experimental results demonstrated that OBSUM outperformed other methods in terms of both accuracy indices and visual effects over time-series. Furthermore, OBSUM also achieved satisfactory results in two typical remote sensing applications. Therefore, it has great potential to generate accurate and high-resolution time-series observations for supporting various remote sensing applications.

## 1. Introduction

Dense satellite image time-series with high spatial resolution can effectively facilitate various remote sensing applications in ecology (Dizhou Guo et al., 2022), agriculture (Gao et al., 2017), and disaster (Zhang et al., 2014). However, no single satellite-based sensor is able to perform fine-scale Earth observation with a daily frequency due to the trade-off between spatial resolution and revisit period. For instance, the Multi-Spectral Instrument (MSI) carried by Sentinel-2 satellite constellation has a relatively long revisit period of 5 days, but a high spatial resolution of 10 m (hereafter, fine image) (Drusch et al., 2012). By contrast, the Ocean and Land Color Instrument (OLCI) carried by Sentinel-3 satellite constellation has a daily revisit frequency, but a coarse spatial resolution of 300 m (hereafter, coarse image) (Donlon et al., 2012). In the past decades, many spatiotemporal fusion methods have been proposed to blend the abovementioned two types of images, thus generating images with both high spatial and temporal resolution (Gao et al., 2015; Ghamisi et al., 2019; Zhu et al., 2018).

Generally, spatiotemporal fusion methods can be divided into five main categories: spatial unmixing-based, weight function-based, Bayesian-based, learning-based, and hybrid methods (Zhu et al., 2018). The Multisensor Multiresolution Technique (MMT) is the first spatial unmixing-based method developed to fuse images with different spatial and temporal resolutions (Zhukov et al., 1999). MMT consists of four steps: (1) classify the input fine image to get the fine-scale land-cover classification map of the base date, (2) calculate the fractions of each land-cover class inside all coarse pixels, (3) unmix the input coarse image using a moving window to get the reflectance of each land-cover class of the prediction date, and (4) assign the unmixed reflectance to the fine-scale land-cover classification map to get the prediction. Based on the MMT framework, many spatial unmixing-based methods have been proposed in the past decades. For example, the Unmixing-Based Data Fusion (UBDF) method applies a constrained least squares method to spatial unmixing to ensure the predicted reflectance are within an appropriate range (Zurita-Milla et al., 2008). UBDF also optimizes the number of classes and the size of the moving window to further improve the fusion accuracy. The Spatial Temporal Data Fusion Approach (STDFA) unmixes the temporal change between two coarse images at both the base and the prediction dates, then adds the unmixed fine-scale temporal changes to the base fine image to get the prediction (Mingquan et al., 2012). In order to remove the block effects (i.e., fine pixels of the same land-cover class inside two adjacent coarse pixels present different reflectance and results in clear footprints of coarse pixels in the predicted image), the Blocks-Removed Spatial Unmixing (SU-BR) method introduces an iterative optimization strategy that considers both residuals of the unmixing model and reflectance differences (Wang et al., 2021). SU-BR can be integrated into any existing spatial unmixing-based method to improve both the fusion accuracy and the visual effect.

The Spatial and Temporal Adaptive Reflectance Fusion Model (STARFM) is the first

weight function-based spatiotemporal fusion method presented in the literature (Gao et al., 2006). STARFM predicts the reflectance of one fine pixel by combining the reflectance change of its spatially neighboring similar pixels. The weights of similar pixels are calculated according to the spectral differences, temporal differences, and spatial distances to the target fine pixel. Many approaches have been proposed to improve the performance of STARFM. The Spatial Temporal Adaptive Algorithm for mapping Reflectance Change (STAARCH) employes a change detection technique to capture both the spatial extent and the temporal evolution of disturbance events, thus improving the robustness to land-cover changes (Hilker et al., 2009). The Enhanced Spatial and Temporal Adaptive Reflectance Fusion Model (ESTARFM) introduces a conversion coefficient to improve the fusion performance in heterogeneous areas (Zhu et al., 2010). There are also some weight function-based methods that are not based on the principle of STRAFM. For example, the three-step method (Fit-FC) first correlates the input coarse images by fitting linear regression models, then applies the model to the input fine image to obtain an initial prediction. After that, similar pixels are employed to refine the prediction by removing the block effects and compensating the residuals. Fit-FC can retrieve strong temporal changes when there is a low correlation between the two input coarse images (Wang & Atkinson, 2018).

Bayesian-based methods consider spatiotemporal fusion as a maximum a posterior (MAP) problem and predict the fine image by optimizing the probability functions (Li et al., 2013; Shen et al., 2016). Learning-based methods employ machine learning techniques, e.g., dictionary pair learning (Huang & Song, 2012), convolutional neural network (Liu et al., 2019) and generative adversarial network (H. Zhang et al., 2021) to model the relationship between the input coarse images and the fine image. A more detailed introduction to Bayesian-based and leaning-based methods can be found in Zhu et al. (2018).

Hybrid methods combines both spatial unmixing technique and the weight function-based methods to improve the prediction robustness. The Flexible Spatiotemporal DAta Fusion (FSDAF) method integrates spatial unmixing, thin plate spline (TPS) interpolation, and weight function to better retrieve land-cover changes. Many optimized versions of FSDAF have been proposed to improve either fusion accuracy or computational efficiency (Gao et al., 2022; Guo et al., 2020). The Reliable and Adaptive Spatiotemporal Data Fusion (RASDF) method introduces a reliability index to describe the sensor differences, optimize the spatial unmixing, and also guide the residual compensation (Shi et al., 2022). The adaptive global unmixing model and adaptive local unmixing model in RASDF are utilized collaboratively to retrieve spectral changes between the coarse images while preserving structural information in the fine image. The Variation-based Spatiotemporal Data Fusion (VSDF) method employs an abundant variation classification (AVC) to detect land-cover changes and guide the spatial unmixing. Moreover, a feature-level edge fusion is introduced to strengthen the edges and texture in the predicted fine image (Xu et al., 2022).

Despite various spatiotemporal fusion methods have been proposed in recent years to improve the prediction accuracy and robustness, there are still several remaining challenges. First, most existing methods are based on pixel-level computation that neglects the object-level information of the land surface. As a result, these methods suffer from both low computational efficiency and intra-class spectral variability (D. Guo et al., 2022). An object is a shape composed of spectrally similar pixels that are spatially adjacent, and all pixels within the object belong to the same land-cover class (Hossain & Chen, 2019). The object-level information is one of the inherent characteristics of the land surface and is valuable for improving the fusion accuracy. In recent years, several object-level fusion methods have been proposed and proved their satisfactory performances, including the Object-Level (OL) weight function methods (D. Guo et al., 2022) and the Object-Based SpatioTemporal Fusion Model (OBSTFM) (Hua Zhang et al., 2021). However, to the best of our knowledge, no existing method has combined object-level information, spatial unmixing, and weight functions to obtain a more accurate fusion result. Secondly, many existing methods have difficulty in accurately retrieving strong temporal changes. For example, the Fit-FC method assumes the temporal changes of the ground objects are scale-invariant, and models the temporal changes by applying local window-based linear regression to the input coarse images. After that, the regression models are applied to the input fine image to obtain an initial prediction. Even if the following spatial filtering step can alleviate the block effect introduced by the regression model fitting, there are still spectral distortions in the final prediction caused by the scale inconsistency, especially in heterogeneous areas (Shi et al., 2022). Thus, it is important to develop a fusion method that can accurately retrieve strong temporal changes while introducing fewer errors and uncertainties.

In order to address the aforementioned problems, an Object-Based Spatial Unmixing Model (OBSUM) is proposed and validated in this paper. OBSUM is a hybrid method that utilizes object-level information to obtain more accurate fusion results. The object-based image analysis (OBIA), spatial unmixing, and combination of similar pixels are integrated into OBSUM. A novel object residual index is proposed to calculate and compensate the residual of each object to improve the fusion accuracy significantly. It is also noteworthy that OBSUM requires a minimum number of input images. The performance of OBSUM is compared to those of five typical spatiotemporal fusion methods, including UBDF, STARFM, Fit-FC, OBSTFM, and FSDAF, using time-series of Sentinel-2 and simulated Sentinel-3 images. The potential of OBSUM to support various remote sensing applications is also discussed in detail.

The remainder of this paper is organized as follows. Section 2 introduces the methodology of OBSUM. Section 3 presents the comparison experiment. Section 4 introduces two potential application scenarios supported by OBSUM. Finally, Section 5 gives a detailed discussion of OBSUM and draws the conclusion on the main findings of this paper.

## 2. Methodology

OBSUM only requires one fine image at the base date $t_b$ and one coarse image at the prediction date $t_p$ to predict the fine image at $t_p$. It is based on four main steps: (1) image preprocessing, (2) object-level unmixing, (3) object-level residual compensation and (4) pixel-level residual compensation. The flowchart of OBSUM is shown in Fig. 1, and the detailed description of the method is given below.

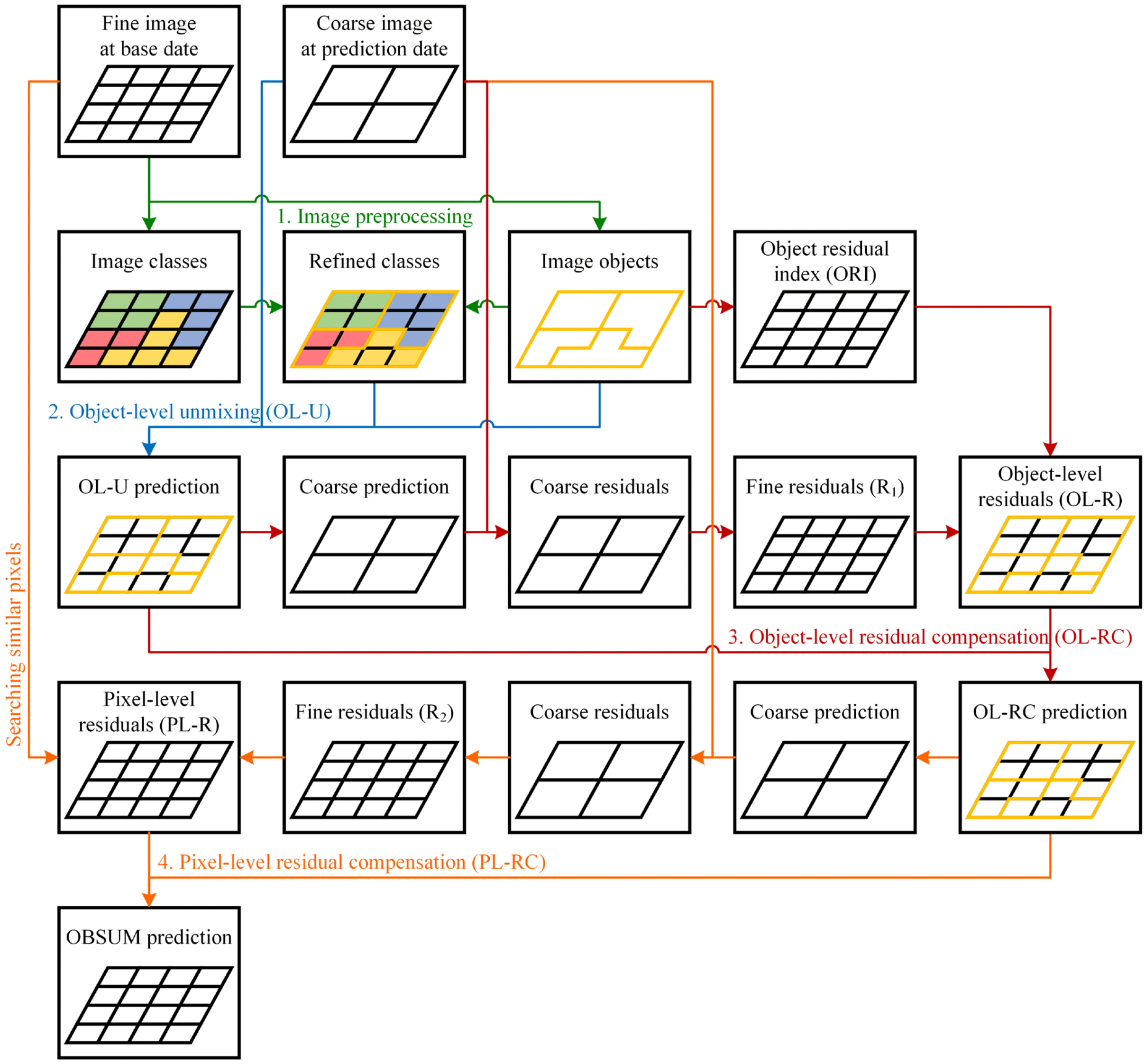

**Fig. 1.** Flowchart of OBSUM. The four steps are represented by connectors with different colors. Images containing the object boundaries (yellow lines) have characteristics similar to the image objects, i.e., all fine pixels within an object have the same value.

### *2.1. Image preprocessing*

The image preprocessing includes: (1) image classification, (2) image segmentation, and (3) refinement of the classification map. First, the fine image at $t_b$ is classified into several land-cover classes. Either a supervised or unsupervised classifier can be applied to obtain an initial classification map, depending on the availability of reference land-cover labels. In this study, we employed the simple unsupervised K-Means classifier (Lloyd, 1982) to make OBSUM fully automatic and independent from a training set.

After image classification, the fine image at $t_b$ is segmented into homogeneous regions, i.e., image objects. In this study, the state-of-the-art segment anything model (SAM) (Kirillov et al., 2023) is used to perform image segmentation. SAM is trained on a dataset that contains 11 million images and 1.1 billion segmentation masks. It can generate masks for all objects in an image and has impressive zero-shot performance on any segmentation task. Specifically, for the input image, given the number of samples in the image and the output normative parameters (including mask filtering threshold, non-maximum suppression, etc.), SAM can get the object mask of all pixels within the image automatically.

After image classification and segmentation, the image objects are used to refine the classification map. The OBIA technique assumes all pixels inside an object have similar spectral characteristics, therefore, these pixels should also belong to the same land-cover class. After classification map refinement, the land-cover class of all fine pixels within and object is set to the mode of the original classes of these fine pixels. As shown in the flowchart in Fig. 1, the graph on row 2 column 1 is the original land-cover classification map, in which there are four land-cover classes represented by different colors. And the graph on row 2 column 3 shows the boundaries of image objects. After classification map refinement, all fine pixels within an object have a uniform land-cover class inside the refined classification map (row 2 column 2 in Fig. 1), which drives the subsequent object-level fusion steps.

### *2.2. Object-level unmixing*

The object-level unmixing (OL-U) incorporates the spatial unmixing technique and OBIA to obtain an initial prediction. According to the spatial unmixing theory, assuming that changes are not scattered within a coarse pixel, the reflectance of the coarse pixel can be modeled as the linear combination of the reflectance of each class of fine pixels within it:

$$C_{tp}(x_i, y_i, b) = \sum_{c=1}^{n_c} f_c(x_i, y_i) \times F_{tp}(c, b) \tag{1}$$

where $C_{tp}(x_i, y_i, b)$ is the band $b$ reflectance of the target coarse pixel at location $(x_i, y_i)$ at $t_p$, $n_c$ is the number of land-cover classes, $f_c(x_i, y_i)$ denotes the fraction of class $c$ inside the target coarse pixel at $t_p$, and $F_{tp}(c, b)$ is the band $b$ reflectance of class $c$ at $t_p$. The spatial unmixing theory assumes the land-cover is stable from $t_b$ to $t_p$, therefore, the refined classification map is used to calculate $f_c(x_i, y_i)$ by counting the proportion of class $c$ inside the coarse pixel.

Inside the target coarse pixel located at $(x_i, y_i)$, $F_{tp}(c, b)$ can be estimated through a local square window with size $w$ centered on this target coarse pixel. The coarse pixels inside this window can compose the following linear equation system:

$$\begin{bmatrix} C_{tp}(x_1, y_1, b) \\ \vdots \\ C_{tp}(x_i, y_i, b) \\ \vdots \\ C_{tp}(x_n, y_n, b) \end{bmatrix} = \begin{bmatrix} f_1(x_1, y_1) & f_2(x_1, y_1) & \cdots & f_{n_c}(x_1, y_1) \\ \vdots & \vdots & & \vdots \\ f_1(x_i, y_i) & f_2(x_i, y_i) & \cdots & f_{n_c}(x_i, y_i) \\ \vdots & \vdots & & \vdots \\ f_1(x_n, y_n) & f_2(x_n, y_n) & \cdots & f_{n_c}(x_n, y_n) \end{bmatrix} \begin{bmatrix} F_{tp}(1, b) \\ \vdots \\ F_{tp}(c, b) \\ \vdots \\ F_{tp}(n_c, b) \end{bmatrix} \tag{2}$$

where $n$ is the number of coarse pixels inside the local window, subject to $n = w^2$. Equation (2) can be solved through the least square method, and the resulting reflectance are assigned to each land-cover class inside each coarse pixel to obtain a preliminary spatial unmixing result.

Following the basic principle of OBIA, the reflectance of the $o$th object is assigned as the mean reflectance of all fine pixels within the object, and the OL-U prediction is obtained by:

$$OL\text{-}U_{tp}(o,b) = \frac{1}{p_o}\sum_{k=1}^{p_o} F_{tp}(k,b) \tag{3}$$

where $OL\text{-}U_{tp}(o,b)$ is the band $b$ reflectance of the $o$th object at $t_p$ predicted by OL-U, $p_o$ is the number of fine pixels within the $o$th object, and $F_{tp}(k,b)$ is the band $b$ reflectance of the $k$th fine pixel within the $o$th object at $t_p$. Fig. 2 shows an example of OL-U. It can be seen that the fine pixels within an object have the same reflectance in the OL-U prediction, which provides a basis for the subsequent object-level residual compensation.

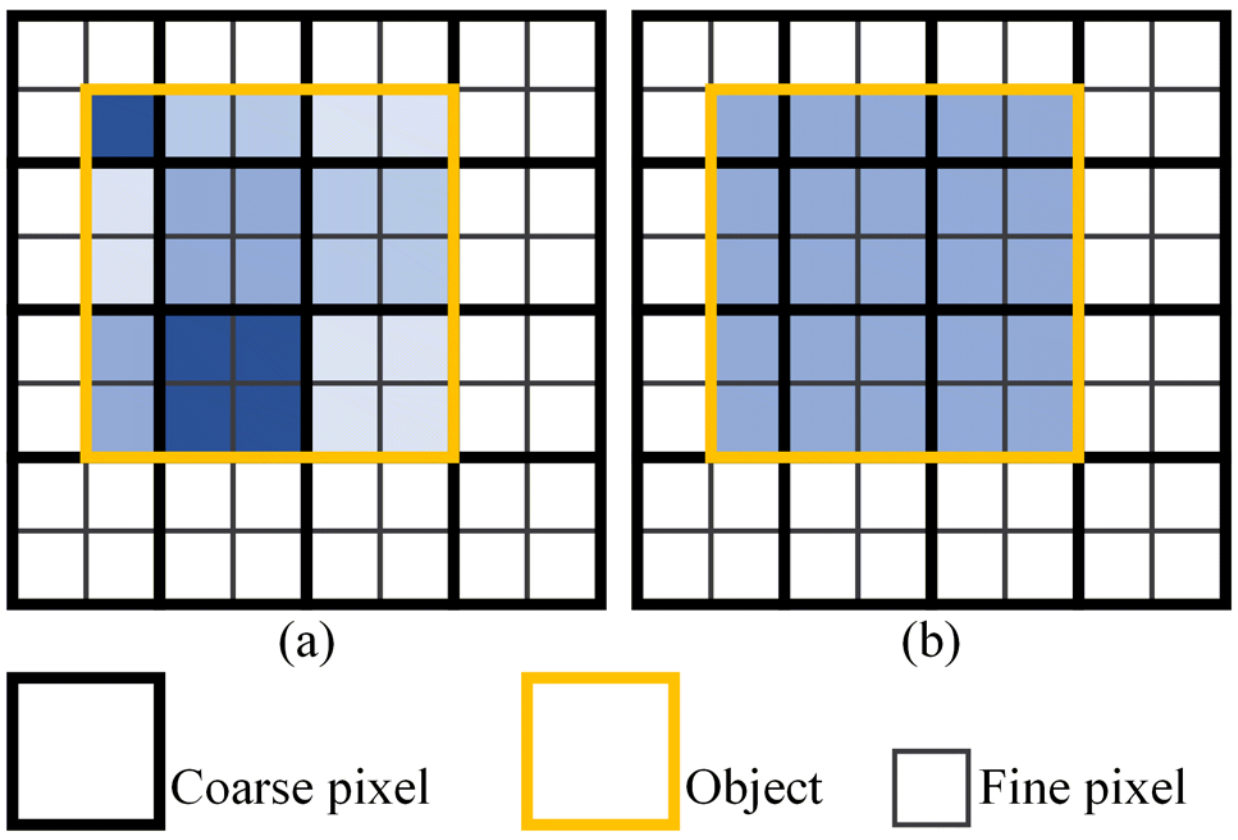

**Fig. 2.** Illustration of OL-U. (a) Result of spatial unmixing, and (b) result of OL-U.

### *2.3. Object-level residual compensation*

The OL-U produces an initial prediction in which all fine pixels within an object have the same reflectance. However, the OL-U prediction is not accurate enough because it assumes the land-cover classes are stable from $t_b$ to $t_p$, while ignoring that the possible presence of changed coarse pixels would decrease the accuracy of the local unmixing process. Thus, residual compensation is necessary to recover the spectral information and improve the prediction accuracy. By assuming all fine pixels within an object have the same residual, the objective of object-level residual compensation (OL-RC) is to calculate and compensate the residual for each object.

The OL-U prediction is first upscaled to the resolution of the coarse image to get a coarse prediction, and the coarse residuals are calculated by subtracting the coarse prediction from the coarse image at $t_p$. As shown in Fig. 3, an object intersects with nine coarse residual pixels, and the residual of the red coarse pixel is most likely to be the residual of this object because this pixel is fully intersected with the object. This is the theoretical basis of the proposed OL-RC. However, considering the complexity and heterogeneity of the land-cover, the coarse residuals are not suitable for residual compensation of small objects, in which no complete coarse residual pixel exists. Therefore, the coarse residuals are downscaled to the resolution of the fine image using a bi-cubic interpolation to obtain the fine residuals $R_1(x_{ij}, y_{ij}, b)$. The scheme of OL-RC needs to be adjusted accordingly.

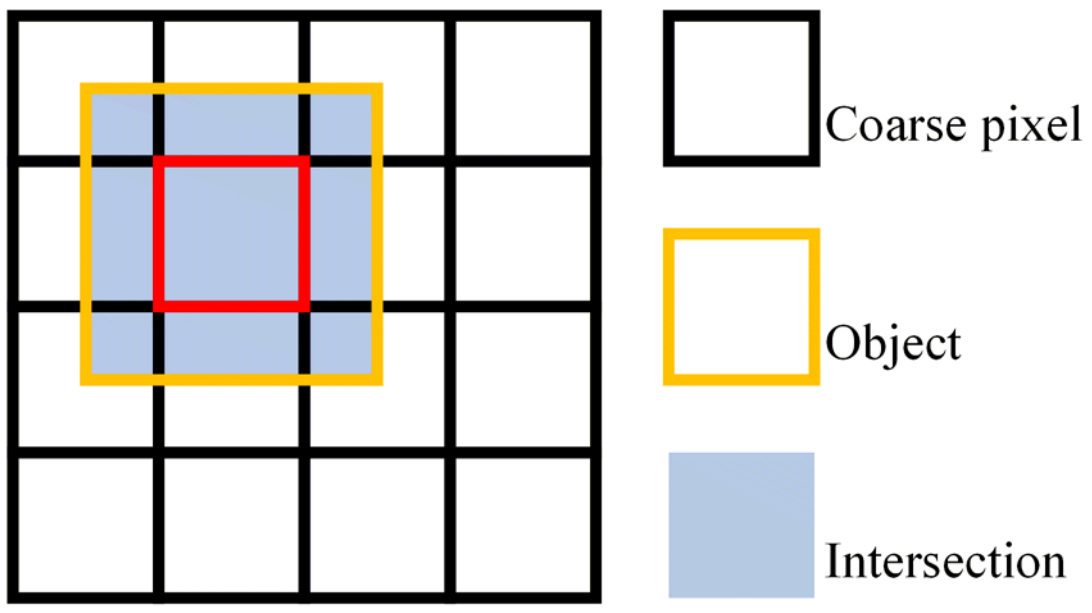

**Fig. 3.** An example of OL-RC using coarse residuals.

Inside the coarse pixel located at $(x_i, y_i)$, the normalized spatial distance $DC$ between the $j$th fine pixel and the center of the coarse pixel is defined as:

$$DC(x_{ij}, y_{ij}) = 1 + \sqrt{\left(x_{ij} - x_{i\frac{m}{2}}\right)^2 - \left(y_{ij} - y_{i\frac{m}{2}}\right)^2} \Big/ (s/2) \quad (4)$$

where $(x_{ij}, y_{ij})$ is the coordinates vector of the $j$th fine pixel, $m$ is the number of fine pixels inside one coarse pixel, $(x_{i\frac{m}{2}}, y_{i\frac{m}{2}})$ is the coordinates vector of the coarse pixel center, and $s$ is the scale factor between the coarse image and the fine image, subject to $m = s^2$. $DC$ ranges between 1 and $1 + \sqrt{2}$, and a lower value indicates a higher similarity between the downscaled fine residual at $(x_{ij}, y_{ij})$ and the original coarse residual at $(x_i, y_i)$.

In order to perform OL-RC, we introduce the object homogeneity index $OHI$:

$$OHI(x_{ij}, y_{ij}) = \frac{1}{m}\sum_{k=1}^{m} I_k \quad (5)$$

When the $k$th fine pixel inside a local window (the window size is one coarse pixel, and the window center is the target fine pixel) belongs to the same object as the target pixel located at $(x_{ij}, y_{ij})$, $I_k = 1$; otherwise, $I_k = 0$. $OHI$ ranges between 0 and 1, and a higher value indicates a more homogeneous object-level land-cover.

Theoretically, if both a fine residual is similar to the residual of the coarse pixel it is located in, and the object-level land-cover inside the coarse pixel is homogeneous, then the fine residual has a higher similarity to the actual residual of the object it is located in. Therefore, $DC$ and $OHI$ are combined to calculate the object residual index ($ORI$):

$$ORI(x_{ij}, y_{ij}) = OHI(x_{ij}, y_{ij})/DC(x_{ij}, y_{ij}) \quad (6)$$

$ORI$ ranges between 0 and 1, and a higher value indicates a higher similarity of the fine residual to the actual object residual.

After calculating $ORI$, a possible scheme for OL-RC is to select a fine pixel with the highest $ORI$ within each object and then assign the fine pixel's residual as the predicted object-level residual. However, given that bicubic interpolation would smooth the residual map and there are uncertainties caused by errors of the previous steps, only one fine residual pixel is not robust enough to compensate the residual of an object. Therefore, we first select a certain percentage of fine residual pixels that have the highest $ORI$ within an object, then combine the selected fine residuals to calculate the object-level residual. The number of selected fine residual pixels for the $o$th object is defined as $r_o$:

$$r_o = OR\ percent \times p_o \quad (7)$$

where $OR\ percent$ is the percentage of selected fine residual pixels for the $o$th object, and $p_o$ is the number of fine pixels inside the $o$th object. Considering the $OR\ percent$, a relatively low value is unstable for calculating the object-level residual, whereas a relatively high value could introduce errors since the bi-cubic interpolation would smooth the residual map. Through trial-and-error experiments, we found the fusion accuracy is stable when $OR\ percent = 15$. Therefore, $OR\ percent$ is set to 15 in the following parts of this paper.

After the selection of fine residual pixels, the weight of the $k$th selected pixel is calculated as:

$$W_k = ORI_k \Big/ \sum_{k=1}^{r_o} ORI_k \tag{8}$$

where $ORI_k$ is the object residual index of the $k$th selected fine pixel.

The predicted object-level residual (OL-R) of the $o$th object is the weighted sum of the residuals of the selected fine pixels:

$$OL\text{-}R(o,b) = \sum_{k=1}^{r} W_k \times R_1(k,o,b) \tag{9}$$

where $R_1(k,o,b)$ is the band $b$ residual of the $k$th selected fine pixel inside the $o$th object.

After that, OL-R is added to the OL-U prediction to get the OL-RC prediction:

$$OL\text{-}RC_{tp}(o,b) = OL\text{-}U_{tp}(o,b) + OL\text{-}R(o,b) \tag{10}$$

### *2.4. Pixel-level residual compensation*

The OL-RC calculates and compensates the residual for each object, and the prediction is much more accurate than that of the OL-U. However, both OL-U and OL-RC assume the land-cover type inside each object is stable from $t_b$ to $t_p$. As a result, these two steps cannot recover the within-object land-cover changes in the predicted image. The pixel-level residual compensation (PL-RC) employs a strategy similar to the weight function-based spatiotemporal fusion methods to recover the within-object land-cover changes and further improve the prediction accuracy.

The OL-RC prediction is first upscaled to the resolution of the coarse image to get a coarse prediction, and the coarse residuals are calculated by subtracting the coarse prediction from the coarse image at $t_p$. After that, the coarse residuals are downscaled to the resolution of the fine image using a bi-cubic interpolation to obtain the fine residuals $R_2(x_{ij}, y_{ij}, b)$. The detailed process of PL-RC is given below.

The spectral distance between a target fine pixel located at $(x_{ij}, y_{ij})$ and its $k$th neighboring fine pixel is defined as:

$$S_k = \frac{1}{n_b} \sum_{b=1}^{n_b} \left| F_{tb}(x_k, y_k, b) - F_{tb}(x_{ij}, y_{ij}, b) \right| \tag{11}$$

where $n_b$ is the number of spectral bands of the fine image, and $F_{tb}(x_k, y_k, b)$ and $F_{tb}(x_{ij}, y_{ij}, b)$ are the band $b$ reflectance of the $k$th neighboring pixel and the target pixel, respectively.

Inside the local square window with size $w_s$ centered on the target fine pixel, a total number of $n_s$ pixels with the smallest $S_k$ are selected as similar pixels. The spatial distance between the target pixel and its $k$th similar pixel is defined as $D_k$:

$$D_k = 1 + \sqrt{(x_k - x_{ij})^2 - (y_k - y_{ij})^2} / (w_s/2) \tag{12}$$

where $D_k$ is a relative distance that ranges between 1 and $1 + \sqrt{2}$.

The pixel-level residual (PL-R) of a target pixel is the weighted combination of the residuals of its neighboring similar pixels. According to Tobler's first law of geography (Tobler, 1970), similar pixels that are spatially closer to the target pixel contribute more to the combined residual than similar pixels that are farther from the target pixel. Therefore, the weight of the $k$th similar pixel is calculated as:

$$W_k = (1/D_k) \Big/ \sum_{k=1}^{n_s} (1/D_k) \quad (13)$$

After that, the residual of the target fine pixel located at $(x_{ij}, y_{ij})$ is calculated as:

$$PL\text{-}R(x_{ij}, y_{ij}, b) = \sum_{k=1}^{n_s} W_k \times R_2(x_k, y_k, b) \quad (14)$$

Finally, the OBSUM prediction is obtained by adding PL-R to the OL-RC prediction:

$$OBSUM_{tp}(x_{ij}, y_{ij}, b) = OL\text{-}RC_{tp}(x_{ij}, y_{ij}, b) + PL\text{-}R(x_{ij}, y_{ij}, b) \quad (15)$$

## 3. Experiment and results

### *3.1. Study sites and dataset*

Sentinel-2 MSI and Sentinel-3 OLCI images are used as fine and coarse images in the experiments. For Sentinel-2 images, we collected the Level-2A atmospherically corrected surface reflectance products that were provided by the European Space Agency. We considered the four 10 m spectral bands of Sentinel-2 images, i.e., blue, green, red, and near infra-red (NIR). In order to avoid the errors introduced by radiometric and geometric inconsistencies between two sensors and thus focus only on the model's performance, we degraded the Sentinel-2 images to a spatial resolution of 300 m to simulate Sentinel-3-like images. This simple strategy, which does not take into account the point spread function of different sensors, was widely used in many spatiotemporal fusion studies (Jiang & Huang, 2022; Wang & Atkinson, 2018; Zhu et al., 2016).

Two agricultural regions were selected as experimental sites to validate the performance of OBSUM. Both sites cover an area of 15 km × 15 km, which corresponds to 1500 × 1500 Sentinel-2 pixels and 50 × 50 Sentinel-3 pixels. For each site, we collected five pairs of cloud-free Sentinel-2 images and simulated the Sentinel-3 images. The first Sentinel-2 image is regarded as the base fine image, and the other four Sentinel-2 images are considered as the images to be fused. The time intervals between the four images to be fused at the two experimental sites are one month and two months, respectively. Moreover, there are strong temporal changes between the base image and the images to be fused.

The first site is located in southwest Butte County, California, United States of America (BC site hereafter), and the major crop type is rice. As shown in Fig. 4 (a), the land-cover is dominated by bare land and water on February 13, 2022. According to the 2022 Crop Progress and Conditions report (USDA, 2022a) released by the United States Department of Agriculture (USDA), the rice in California is planted in May, emerges in June, matures in July and August, and is finally harvested in October. The phenology of rice can be clearly observed in Fig. 4 (b)-(e), so the BC site is considered a representative region of phenological changes.

The second site is located in the Coleambally Irrigation Area, New South Wales, Australia (CIA site hereafter), and has a mixed planting of rice, cotton, and corn (Emelyanova et al., 2013). As shown in Fig. 5 (a)-(e), the CIA site has an uneven trend of land-cover changes because of its complex crop planting structure. Moreover, cropland parcels in the CIA site are more irregular-shaped and smaller than those in the BC site. Therefore, the CIA site is considered a heterogeneous region with rapid land-cover changes.

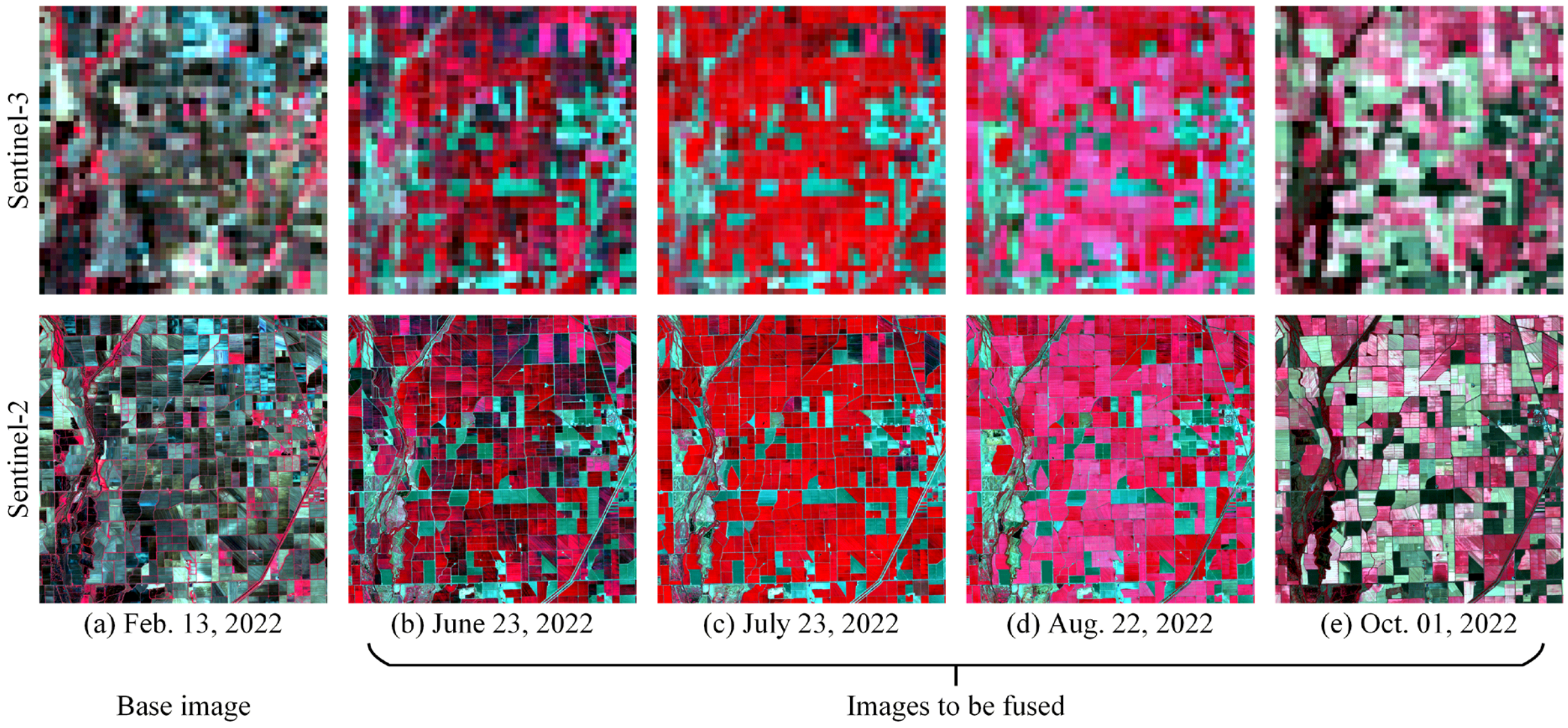

**Fig. 4.** Experimental data of the BC site. (a) Sentinel-2 and simulated Sentinel-3 image pair acquired on Feb. 13, 2022, (b)-(e) four Sentinel-2 and simulated Sentinel-3 image pairs acquired on June 23, 2022, July 23, 2022, Aug. 22, 2022, and Oct. 01, 2022, respectively. All images use NIR-red-green as RGB.

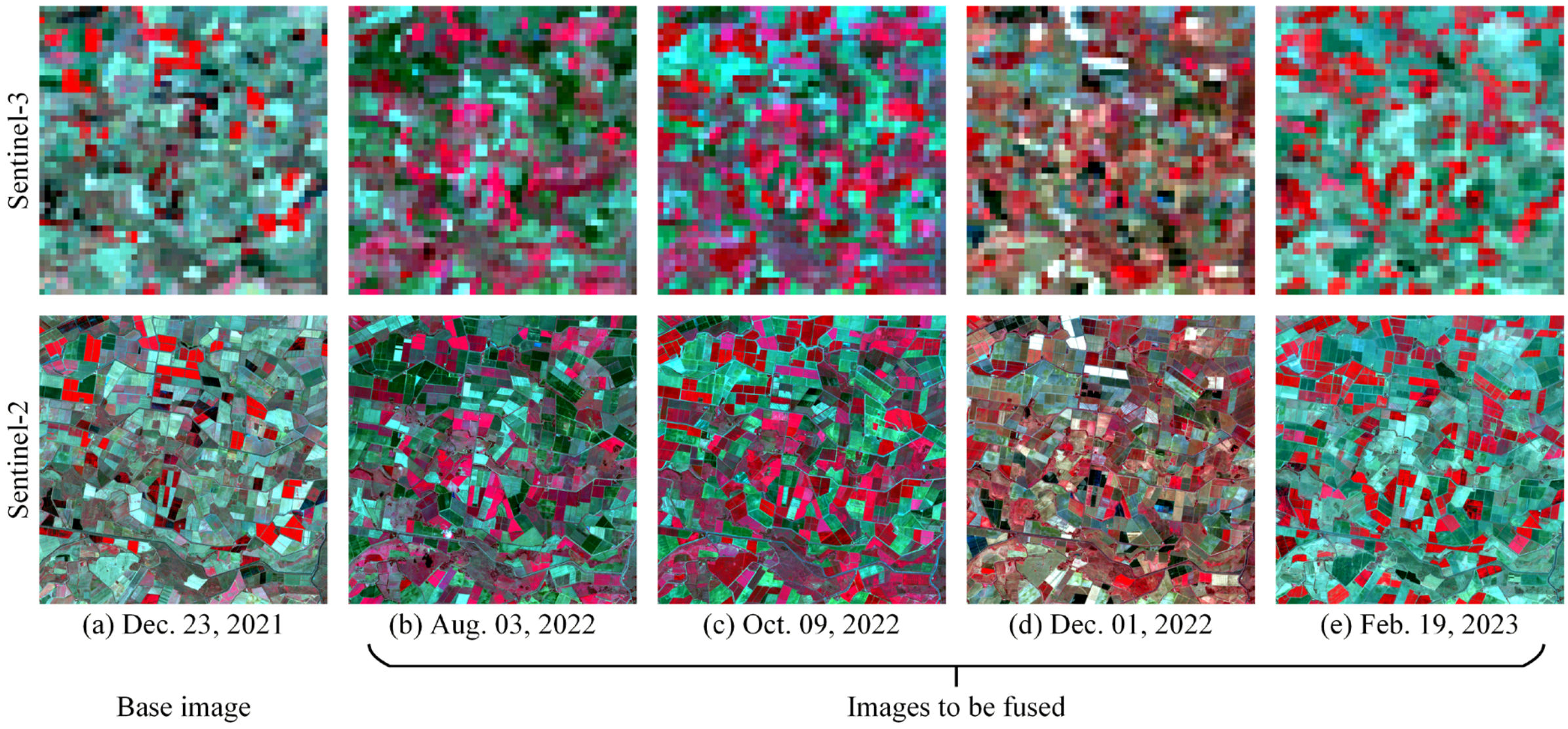

**Fig. 5.** Experimental data of the CIA site. (a) Sentinel-2 and simulated Sentinel-3 image pair acquired on Dec. 23, 2021, (b)-(e) four Sentinel-2 and simulated Sentinel-3 image pairs acquired on Aug. 03, 2022, Oct. 09, 2022, Dec. 01, 2022, and Feb. 19, 2022, respectively. All images use NIR-red-green as RGB.

### *3.2. Experimental setup*

Five spatiotemporal fusion methods that require one pair of fine and coarse images are selected to compare with OBSUM, including UBDF, STARFM, Fit-FC, OBSTFM, and FSDAF. These five methods belong to different categories: UBDF is an unmixing-based method, STARFM and Fit-FC are weight function-based methods, and OBSTFM and FSDAF are hybrid methods. Table 1 shows the parameter settings of the comparison methods and OBSUM, including the number of land-cover classes ($n_c$), the size of the local moving window ($w$), the size of the window for similar pixel selection ($w_s$), and the number of similar pixels ($n_s$). Please note that in Fit-FC, $w$ refers to the size of the local window for regression model fitting (RM); while in UBDF and OBSUM, $w$ represents the size of the local window for spatial unmixing.

**Table 1** Parameter settings for five comparison methods and OBSUM.

| Method | $n_c$ | $w$ | $w_s$ | $n_s$ |
|---|---|---|---|---|
| UBDF | 5 | 11 | N/A | N/A |
| STARFM | 5 | N/A | 31 | 30 |
| Fit-FC | N/A | 3 | 31 | 30 |
| OBSTFM | N/A | N/A | N/A | N/A |
| FSDAF | 5 | N/A | 31 | 30 |
| OBSUM | 5 | 11 | 31 | 30 |

The parameter settings of Fit-FC are the same as in the original research paper (Wang & Atkinson, 2018), and the definition of $w_s$ and $n_s$ values for other methods follows those of Fit-FC. Through visual interpretation of the Sentinel-2 images, the number of land-cover classes ($n_c$) is set to 5 for both sites. Both UBDF and OBSUM contain local window-based spatial unmixing. A small window is not robust enough for local unmixing and would cause clear block effects in the fused image (Wang et al., 2021). Through trial-and-error experiments, we found that a large window could obtain a more accurate local unmixing result. Therefore, the size of the local unmixing window ($w$) is set to 11 in UBDF and OBSUM.

Four indices are used for quantitative evaluation of the fusion accuracy, including average difference (AD), root mean squared error (RMSE), correlation coefficient ($r$), and structural similarity (SSIM) (Zhou et al., 2004). AD ranges from $-1$ to 1, and the optimal value is 0. A positive AD value indicates overestimation of the temporal change between the based image and the image to be fused, whereas a negative AD value indicates underestimation of the temporal change between the two images (Zhu et al., 2022). RMSE ranges between 0 and 1, and a lower value indicates higher fusion accuracy. Both $r$ and SSIM range between 0 and 1, and a larger value indicates higher fusion accuracy.

### *3.3. Comparison of the three fusion steps of OBSUM*

Fig. 6 and Fig. 7 show the base image, the results of the three fusion steps of OBSUM, and the reference image when predicting the fine image at the BC site on June 23, 2022 and the fine image at the CIA site on Aug. 03, 2022, respectively. In both figures, the first row shows the complete images, while the second row shows the enlarged sub-area marked in the yellow rectangle in the first row of sub-figure (e). One can see from Fig. 6 (b) and Fig. 7 (b) that there is no block effect in the images predicted by OL-U, whereas there are several spectral distortions in the OL-U predictions when compared to the reference image. These spectral distortions are mainly introduced by errors in the local unmixing process and land-cover changes between $t_b$ and $t_p$. The following OL-RC step improved the fusion accuracy through computing and compensating the residual for each object. As shown in Fig. 6 (c) and Fig. 7 (c), the OL-RC predictions are visually closer to the reference images, and the spectral information in the yellow dashed ellipses is recovered properly. However, the OL-RC is an object-level processing, which neglects the pixel-level spectral details and the within-object land-cover changes. After the final pixel-level residual compensation step, the OBSUM predictions are more similar to the reference images because PL-RC recovered the pixel-level spectral details and the within-object land-cover changes properly. As shown in Fig. 6 (d), the object in the yellow dashed ellipse is brighter than that in the OL-RC prediction and is visually closer to the reference image. In Fig. 7 (d), the within-object land-cover change in the yellow dashed ellipse is recovered by PL-RC in the final OBSUM prediction.

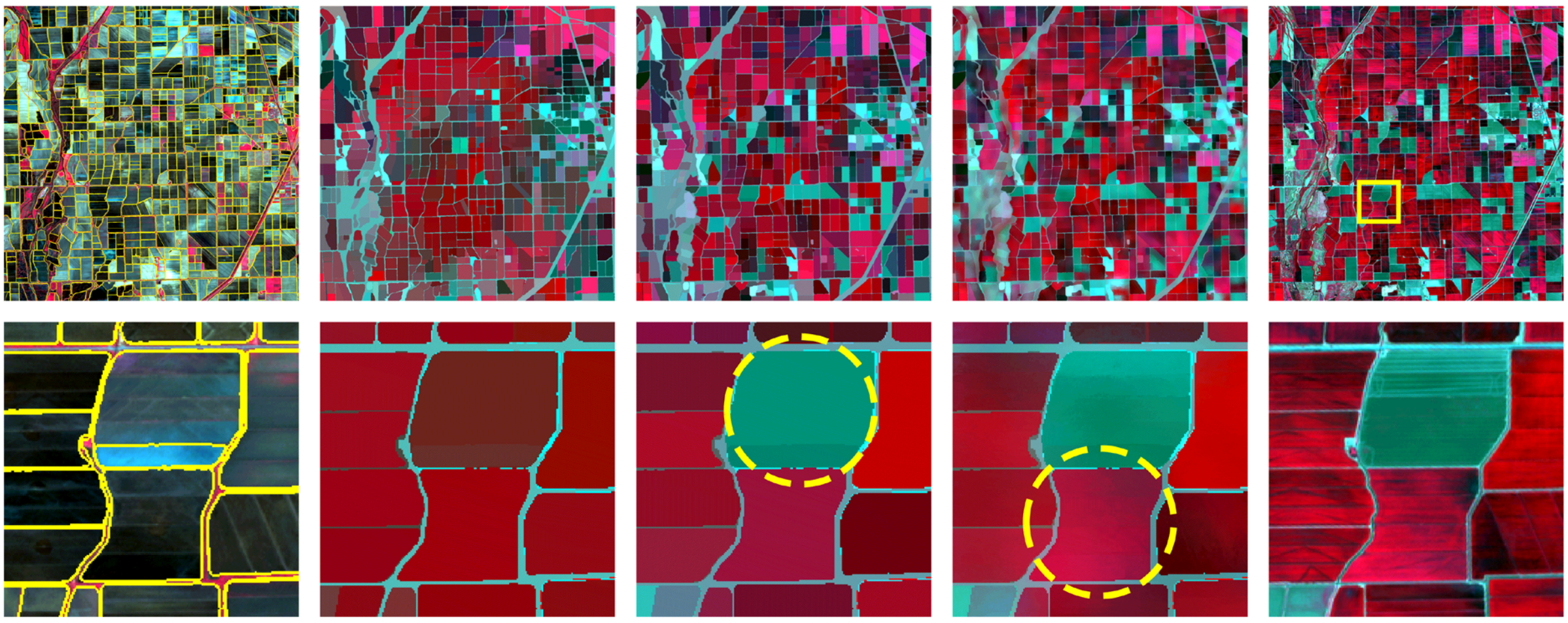

(a) Feb. 13, 2022 (b) OL-U (c) OL-RC (d) OBSUM (e) June 23, 2022

**Fig. 6.** Results of the three different steps of OBSUM at the BC site on June 23, 2022. (a) Base image acquired on Feb. 12, 2022 and the referred segmentation result, (b) OL-U, (c) OL-RC, (d) OBSUM, and (e) reference image. The second row shows the zoomed-in results of the sub-area marked in the yellow rectangle in the upper right sub-figure. The ellipses represented by the yellow dashed line show the recovery of spectral information by two residual compensation steps. All images use NIR-red-green as RGB.

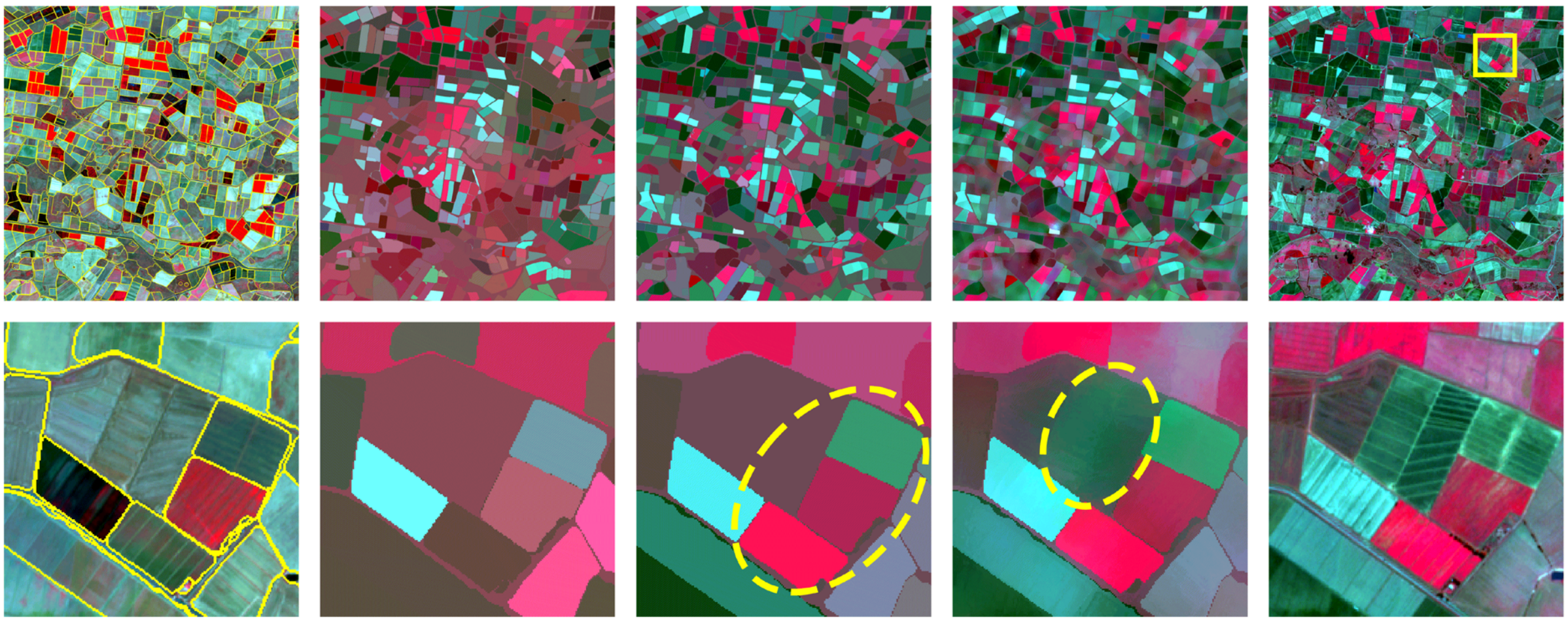

(a) Dec. 23, 2021 (b) OL-U (c) OL-RC (d) OBSUM (e) Aug. 03, 2022

**Fig. 7.** Results of the three different steps of OBSUM at the CIA site on Aug. 03, 2022. (a) Base image acquired on Dec. 23, 2021 and the referred segmentation result, (b) OL-U, (c) OL-RC, (d) OBSUM, and (e) reference image. The second row shows the zoomed-in results of the sub-area marked in the yellow rectangle in the upper right sub-figure. The ellipses represented by the yellow dashed line show the recovery of spectral information by two residual compensation steps. All images use NIR-red-green as RGB.

Table 2 shows the accuracies of the three fusion steps of OBSUM and the accuracy gains over the previous step when predicting all images to be fused. One can see that the fusion accuracies are gradually improved through the OL-RC and PL-RC steps. For example, when predicting the fine image at the BC site on June 23, 2022, the RMSE decreased by 0.01383 from OL-U to OL-RC, and then decreased by 0.00367 from OL-RC to OBSUM. When predicting the fine image at the CIA site on Aug. 03, 2022, the $r$ increased by 0.26048 from OL-U to OL-RC, and then increased by 0.04513 from OL-RC to OBSUM. It can also be noticed that the accuracies improved by OL-RC (gain over OL-U) are much higher than those improved by PL-RC (gain over OL-RC). For example, the RMSE reduction obtained by OL-RC (0.01831) is approximately four times more than that obtained by PL-RC (0.00382) when predicting the fine image at the BC site on July 23, 2022. Such a behavior can be explained by the fact that the OL-RC can improve the spectral accuracy for almost all objects in the image, whereas PL-RC is designed only for recovering within-object land-cover change, which is relatively rare. Both the visual and quantitative evaluations indicate that all three fusion steps in OBSUM are indispensable, and the two residual compensation steps can recover object-level spectral information and within-object land-cover changes, respectively. Further discussion of the proposed OL-RC process will be given in Section 4.1.

**Table 2** Accuracy of the three fusion steps of OBSUM and accuracy gains over the previous step. The bold values indicate the highest accuracy in each term.

| Site | $t_p$ | Metric | OL-U | OL-RC (gain over OL-U) | OBSUM (gain over OL-RC) |
|---|---|---|---|---|---|
| BC | June 23, 2022 | AD | 0.00017 | 0.00035 (-0.00017) | **0.00001** (0.00034) |
| | | RMSE | 0.04438 | 0.03055 (0.01383) | **0.02688** (0.00367) |
| | | $r$ | 0.52816 | 0.79770 (0.26953) | **0.84426** (0.04656) |
| | | SSIM | 0.86522 | 0.90536 (0.04014) | **0.91061** (0.00525) |
| | July 23, 2022 | AD | 0.00023 | -0.00003 (0.00020) | **0.00000** (0.00002) |
| | | RMSE | 0.05026 | 0.03196 (0.01831) | **0.02776** (0.00420) |
| | | $r$ | 0.56405 | 0.83666 (0.27261) | **0.87806** (0.04139) |
| | | SSIM | 0.82428 | 0.90139 (0.07712) | **0.90728** (0.00588) |
| | Aug. 22, 2022 | AD | 0.00024 | -0.00004 (0.00019) | **0.00001** (0.00003) |
| | | RMSE | 0.04533 | 0.02850 (0.01683) | **0.02469** (0.00382) |
| | | $r$ | 0.50199 | 0.80789 (0.30589) | **0.85446** (0.04657) |
| | | SSIM | 0.88889 | 0.92926 (0.04038) | **0.93400** (0.00474) |
| | Oct. 01, 2022 | AD | 0.00054 | -0.00084 (-0.00030) | **-0.00002** (0.00082) |
| | | RMSE | 0.06548 | 0.03821 (0.02727) | **0.03335** (0.00486) |
| | | $r$ | 0.52753 | 0.86079 (0.33326) | **0.89475** (0.03396) |
| | | SSIM | 0.83066 | 0.89143 (0.06077) | **0.89880** (0.00737) |
| CIA | Aug. 03, 2022 | AD | -0.00009 | 0.00039 (-0.00030) | **-0.00000** (0.00039) |
| | | RMSE | 0.03374 | 0.02165 (0.01208) | **0.01880** (0.00286) |
| | | $r$ | 0.54587 | 0.80635 (0.26048) | **0.85148** (0.04513) |
| | | SSIM | 0.91799 | 0.94172 (0.02373) | **0.94616** (0.00445) |
| | Oct. 09, 2022 | AD | -0.00003 | -0.00027 (-0.00024) | **-0.00002** (0.00025) |
| | | RMSE | 0.04984 | 0.03225 (0.01759) | **0.02831** (0.00394) |
| | | $r$ | 0.45542 | 0.76483 (0.30941) | **0.81626** (0.05143) |
| | | SSIM | 0.84111 | 0.89472 (0.05361) | **0.90179** (0.00707) |
| | Dec. 01, 2022 | AD | 0.00004 | 0.00042 (-0.00038) | **-0.00000** (0.00042) |
| | | RMSE | 0.03461 | 0.02326 (0.01135) | **0.02030** (0.00296) |
| | | $r$ | 0.42288 | 0.78440 (0.36152) | **0.84024** (0.05584) |
| | | SSIM | 0.91931 | 0.93886 (0.01954) | **0.94329** (0.00443) |
| | Feb. 19, 2023 | AD | -0.00007 | 0.00012 (-0.00005) | **-0.00000** (0.00012) |
| | | RMSE | 0.04060 | 0.02310 (0.01749) | **0.02029** (0.00281) |
| | | $r$ | 0.49896 | 0.86390 (0.36493) | **0.89763** (0.03373) |
| | | SSIM | 0.90009 | 0.93917 (0.03908) | **0.94355** (0.00438) |

### *3.4. Comparison with other methods*

Fig. 8 and Fig. 9 show the fusion results of five comparison methods and OBSUM, and the reference images on four prediction dates at the BC site and the CIA site, respectively. For convenience, only the fusion results on the first prediction date are used to perform the visual comparison, i.e., sub-figure (a) in Fig. 8 and Fig. 9.

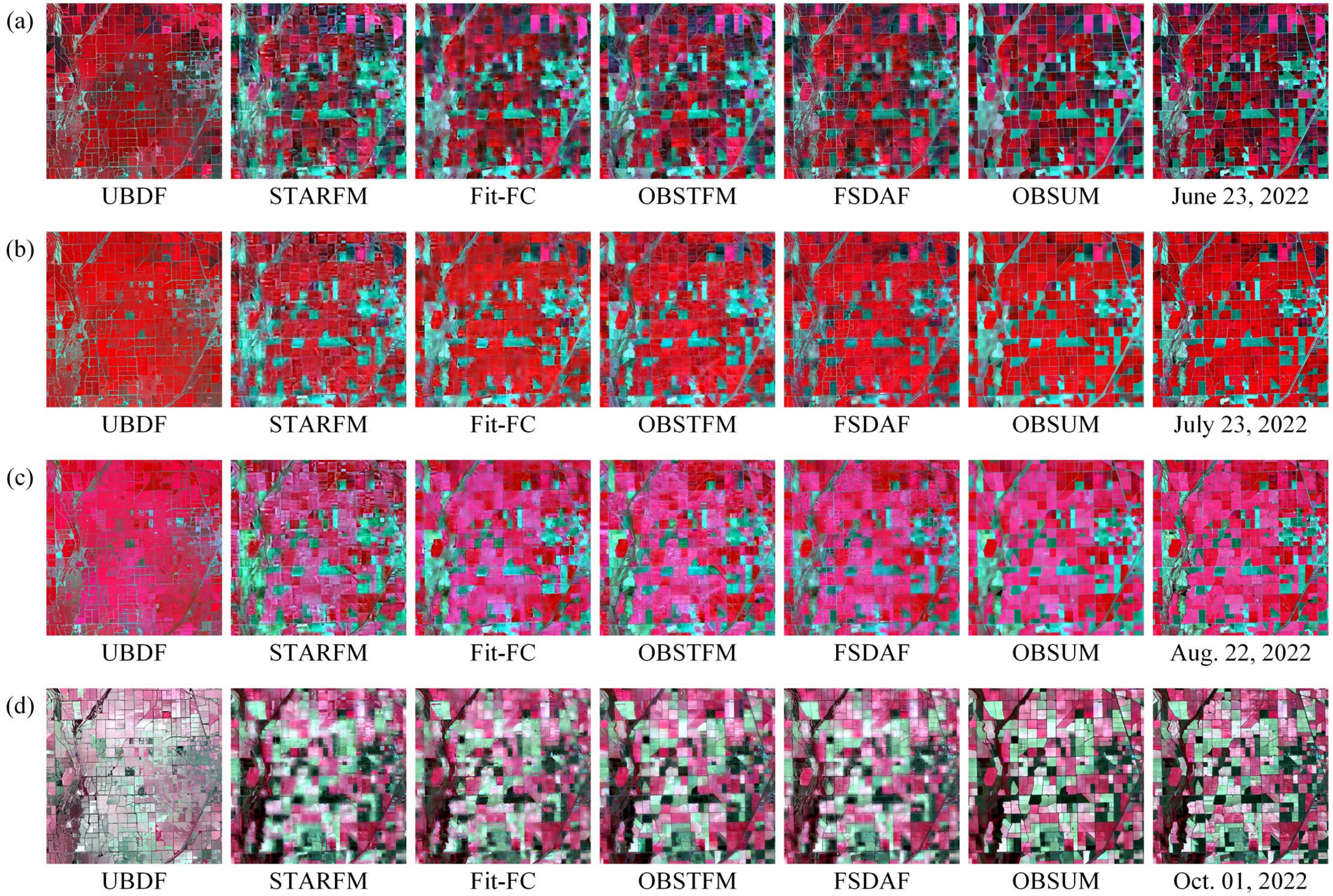

**Fig. 8.** Fusion results of five comparison methods and OBSUM on four prediction dates at the BC site. All images use NIR-red-green as RGB.

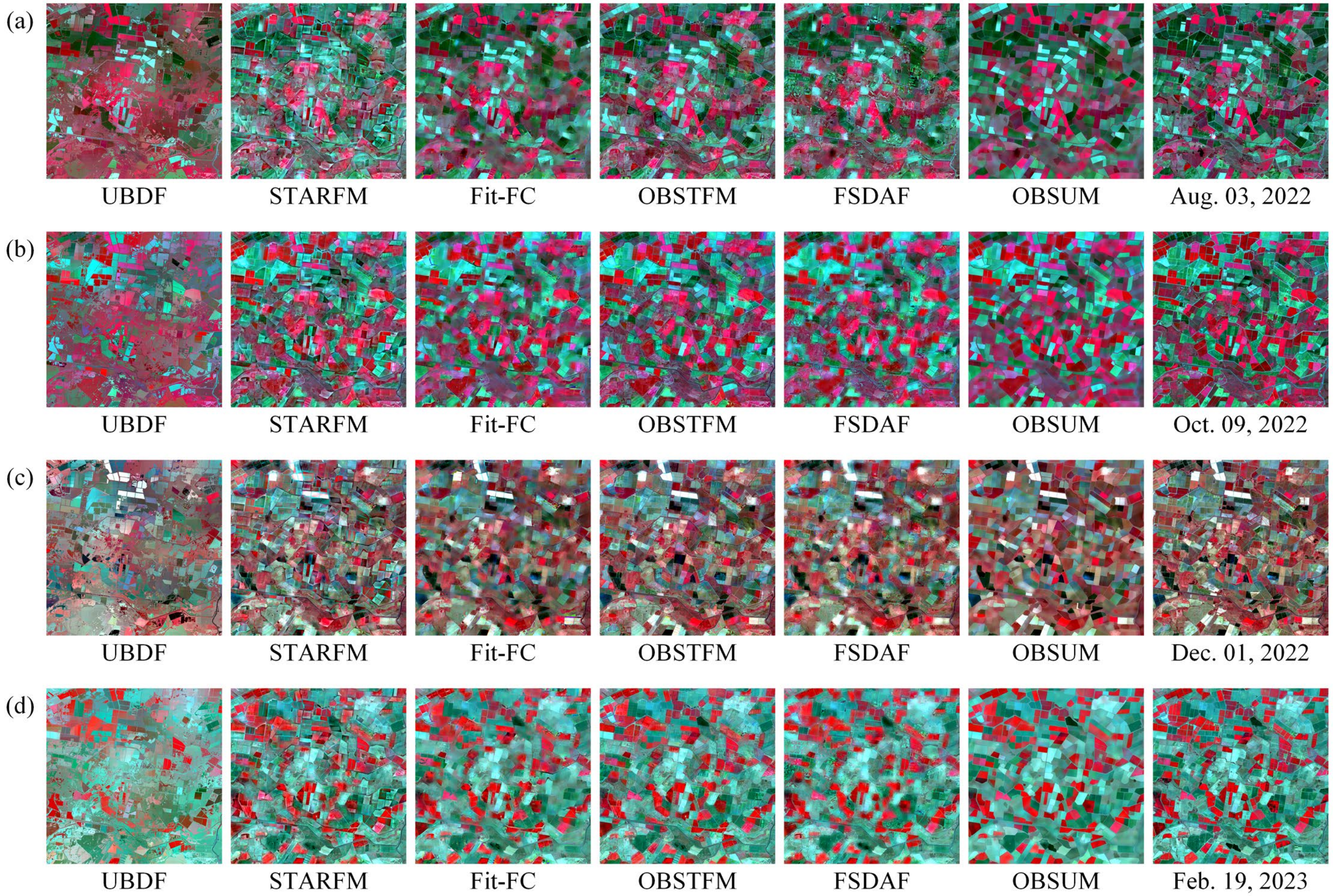

**Fig. 9.** Fusion results of five comparison methods and OBSUM on four prediction dates at the CIA site. All images use NIR-red-green as RGB.

Fig. 10 shows the results of different spatiotemporal fusion methods at the BC site on the first prediction date, and Fig. 11 shows the enlarged sub-areas marked in the yellow rectangle in Fig. 10 (e). One can observe from Fig. 10 that almost all methods predicted a fused image similar to the reference image and captured the strong temporal change between Feb. 13, 2022 and June 23, 2022, except UBDF, which failed to accurately predict the bare lands (light gray objects) in the fused image. In Fig. 11 (b) and (f), there are spectral distortions with the footprints of the classified fine pixels in the predictions of both UBDF and FSDAF (see the dark red pixels in the top left yellow dashed ellipses). This is because the spatial unmixing step in these two methods directly assigns the inaccurate unmixed reflectance according to the fine classification image. UBDF also wrongly estimated the temporal change located inside the bottom right ellipse in Fig. 11 (b). Taking advantage of the classification map refinement and the object-level unmixing, these spectral distortions are not present in the OBSUM prediction. One can also notice that both STARFM and OBSTFM failed to accurately predict the temporal change of the croplands (see the gray and white pixels in the yellow dashed ellipses), and there is an obvious spectral distortion in the Fit-FC prediction (see the light red pixels in the yellow dashed ellipses). By contrast, OBSUM successfully retrieved the temporal change and obtained the highest visual accuracy. The reason is that OBSUM compensates the residuals at both the object-level and pixel-level, and these two residual compensation steps can capture strong temporal change and recover the spectral details properly.

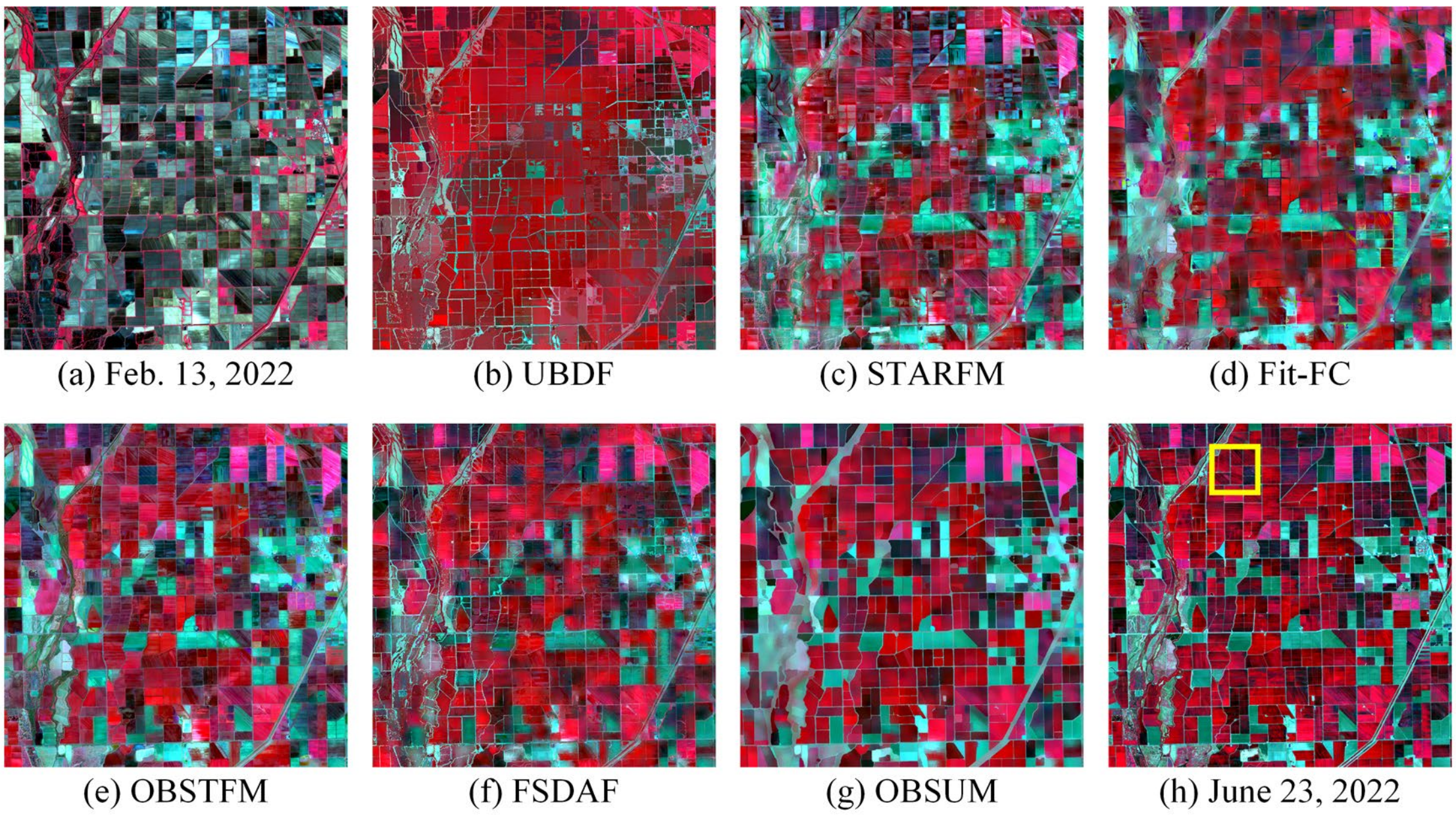

**Fig. 10.** Results of different spatiotemporal fusion methods at the BC site on June 23, 2022. (a) Base image acquired on Feb. 12, 2022, (b)-(g) fusion results of UBDF, STARFM, Fit-FC, OBSTFM, FSDAF, and OBSUM, and (h) reference image. All images use NIR-red-green as RGB.

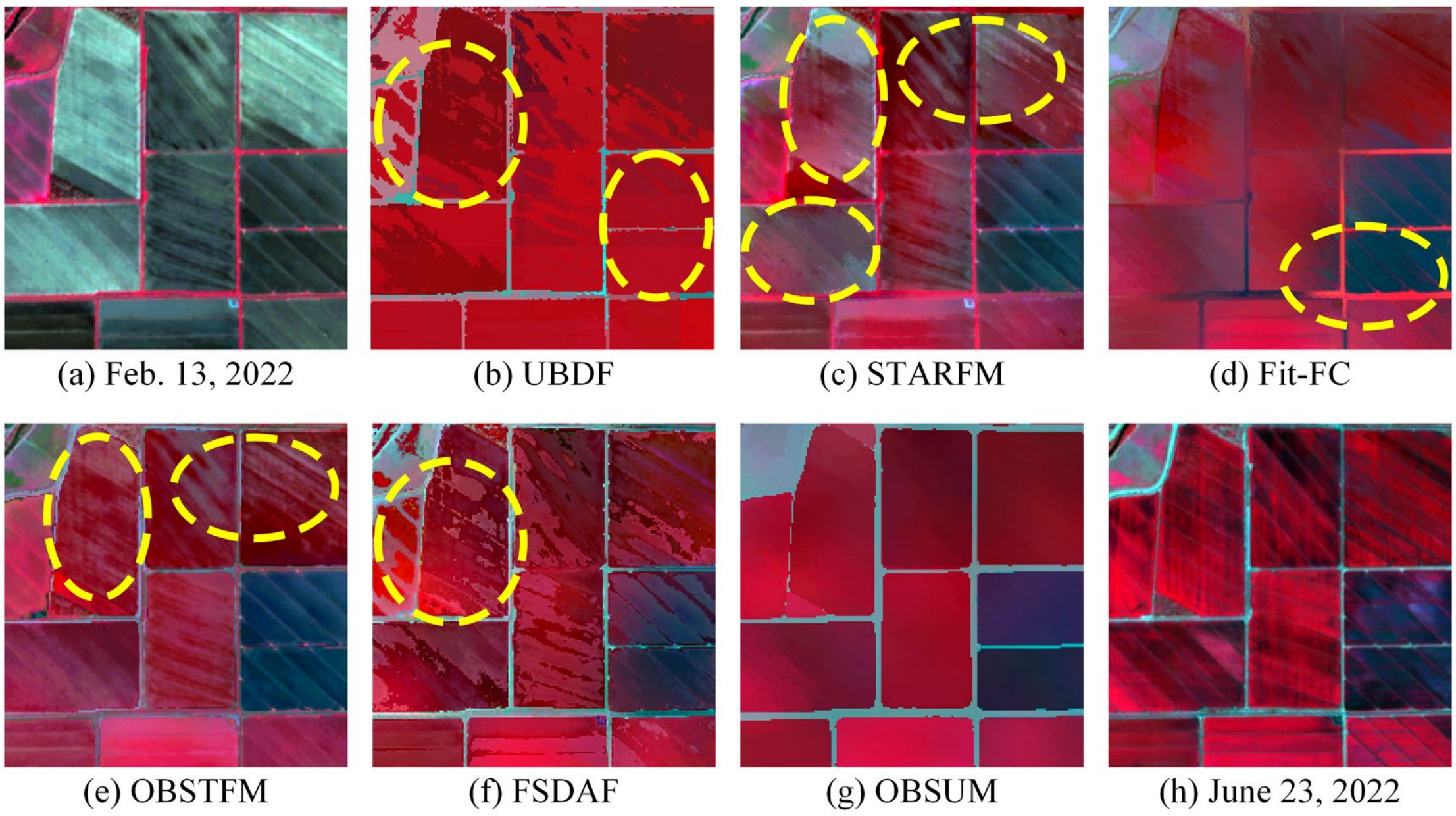

**Fig. 11.** Zoomed-in results of the sub-area marked in the yellow rectangle in Fig. 10 (h). The ellipses represented by the yellow dashed line show the spectral distortion in the fused images. All images use NIR-red-green as RGB.

Fig. 12 shows the results of different spatiotemporal fusion methods at the CIA site on the first prediction date, while Fig. 13 shows the enlarged sub-areas marked in the yellow rectangle in Fig. 12 (e). One can observe from Fig. 12 (b) that UBDF overestimated the extent of croplands, and the prediction is inaccurate. The complete fused images of the other five methods are similar to the reference image. However, as shown in Fig. 13, UBDF and STARFM failed to capture the land-cover change of the narrow croplands marked in the yellow dashed ellipses, while OBSTFM and FSDAF underestimated the temporal change of the crops (see the color difference between the objects in the yellow dashed ellipses and the corresponding objects in the reference image). Moreover, there are many block effects in the UBDF and STARFM predictions in Fig. 13 (b) and (c). In Fig. 13 (d), the Fit-FC prediction in the bottom left yellow dashed ellipse is darker than its counterpart in the reference image, which indicates that the temporal change is underestimated. In addition, there is also spectral distortion in the prediction caused by the regression model fitting step in Fit-FC (see the orange pixels in the bottom right yellow dashed ellipse in Fig. 13 (d)). Taking advantage of object-level processing and two residual compensation steps, only OBSUM accurately predicted the land-cover change in the fused image and obtained the most accurate prediction among these six spatiotemporal fusion methods. This can be demonstrated by the prediction of the narrow pink object and the narrow dark red object at the bottom left of Fig. 13 (g).

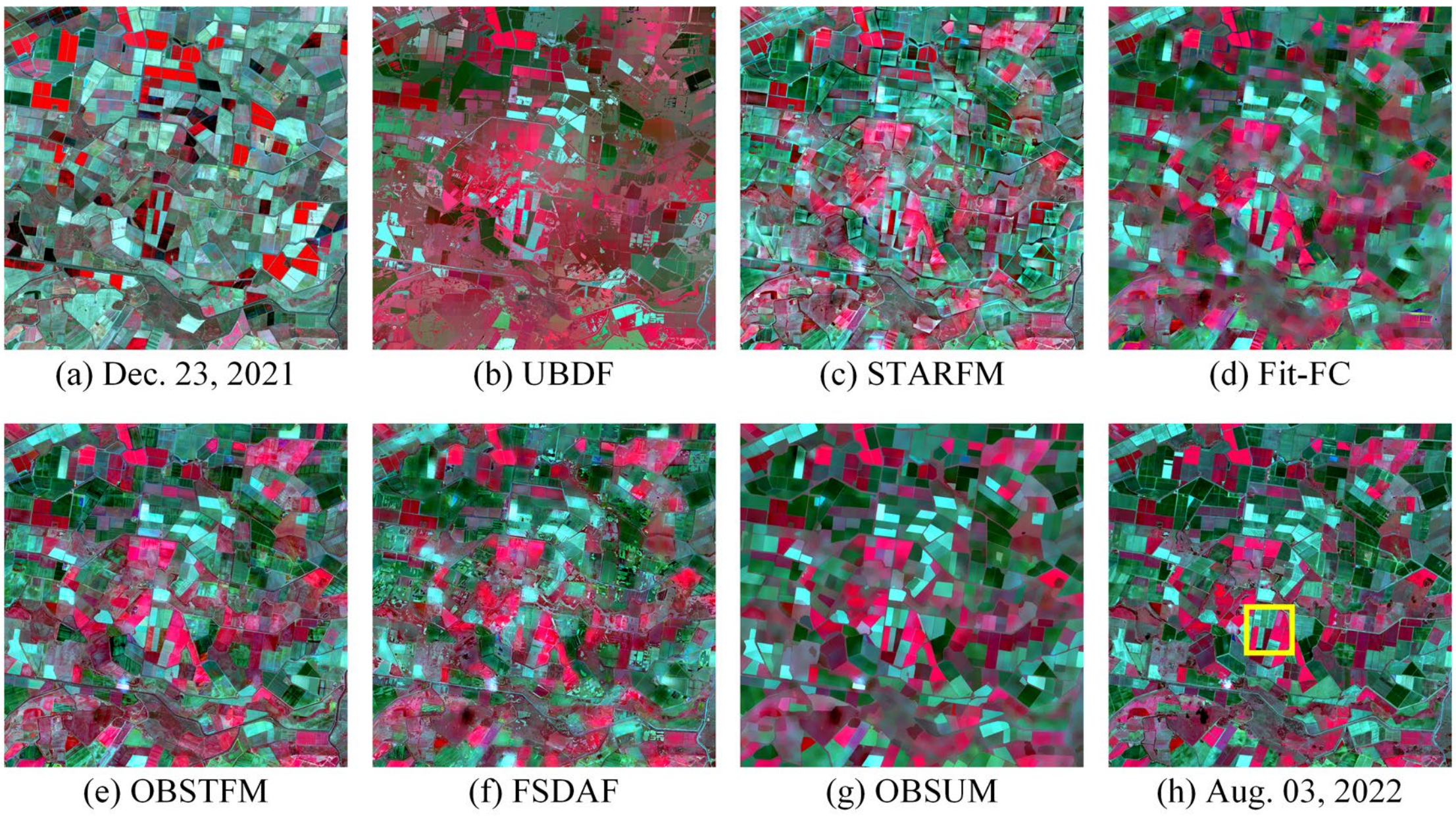

**Fig. 12.** Results of different spatiotemporal fusion methods at the CIA site on Aug. 03, 2022. (a) Base image acquired on Dec. 23, 2021, (b)-(g) fusion results of UBDF, STARFM, Fit-FC, OBSTFM, FSDAF, and OBSUM, and (h) reference image. All images use NIR-red-green as RGB.

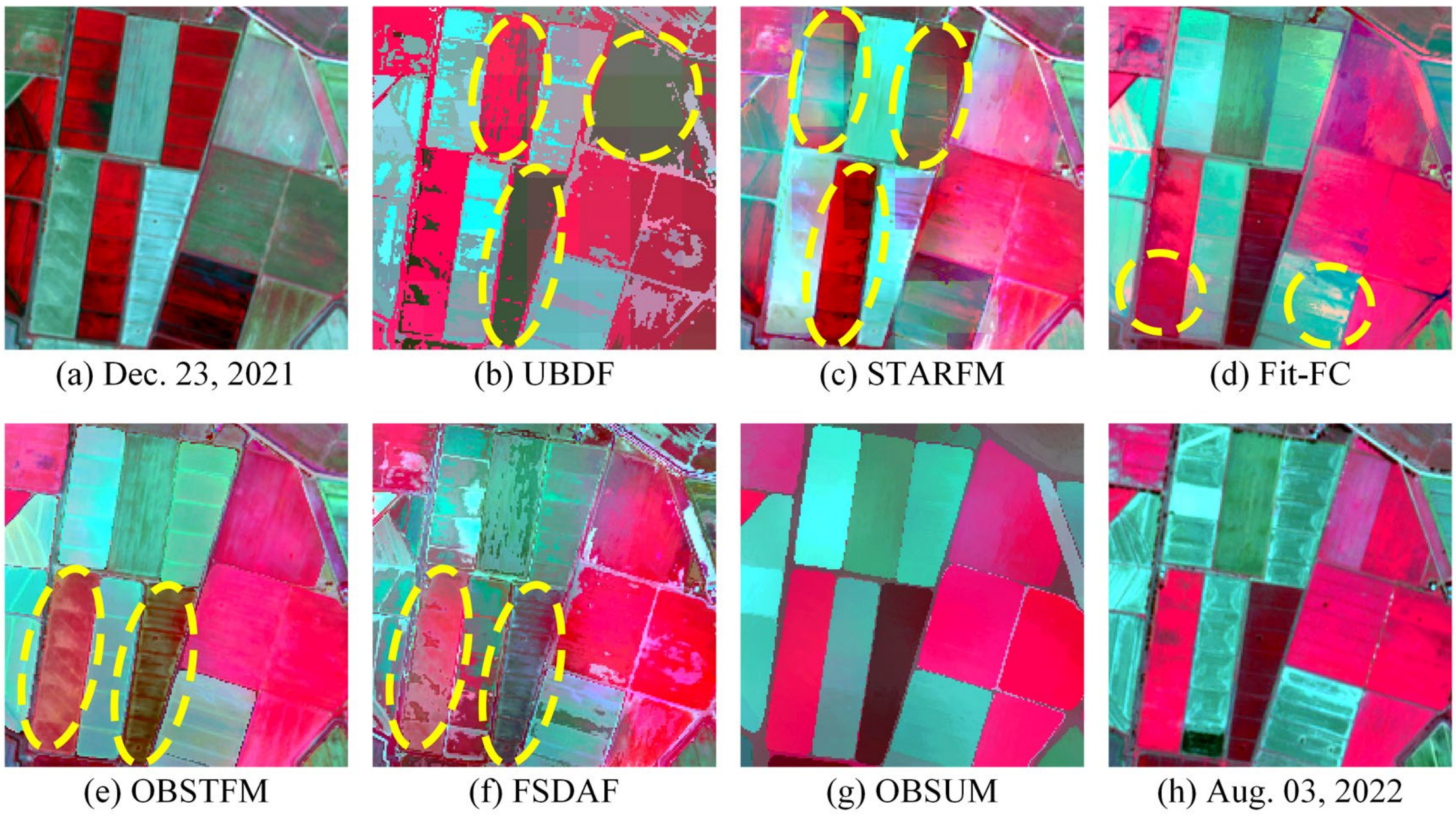

**Fig. 13.** Zoomed-in results of the sub-area marked in the yellow rectangle in Fig. 12 (h). The ellipses represented by the yellow dashed line show the spectral distortion in the fused images. All images use NIR-red-green as RGB

Table 3 gives the accuracies of different spatiotemporal fusion methods on four prediction dates at both sites and the mean accuracy over the time-series. Since a positive and a negative AD at two prediction dates could cancel out each other, the mean AD value over the time-series is not reported for method comparison. One can observe in Table 3 that OBSUM outperformed the comparison methods in terms of four accuracy indices in most fusion tasks, and also obtained the highest average fusion accuracy over the time-series. In the BC site, only OBSUM achieved a mean RMSE less than 0.03, a mean $r$ larger than 0.85, and a mean SSIM larger than 0.90. In the CIA site, only OBSUM achieved a mean $r$ larger than 0.80, and the mean RMSE and mean SSIM of OBSUM are superior to those of the other methods. Both the visual and quantitative evaluations indicate that OBSUM can retrieve strong temporal changes and recover land-cover changes of ground objects, thereby obtaining the highest fusion accuracy in the comparison.

**Table 3** Accuracy of different spatiotemporal fusion methods. The bold values indicate the highest accuracy in each term.

| Site | $t_p$ | Metric | Method | | | | | |
|---|---|---|---|---|---|---|---|---|
| | | | UBDF | STARFM | Fit-FC | OBSTFM | FSDAF | OBSUM |
| BC | June 23, 2022 | AD | 0.00008 | -0.00011 | -0.00002 | -0.00011 | 0.00025 | **0.00001** |
| | | RMSE | 0.05069 | 0.04019 | 0.03540 | 0.03300 | 0.03353 | **0.02688** |
| | | $r$ | 0.45274 | 0.63882 | 0.74044 | 0.75614 | 0.76159 | **0.84426** |
| | | SSIM | 0.82073 | 0.83522 | 0.86998 | 0.88193 | 0.86922 | **0.91061** |
| | July 23, 2022 | AD | 0.00011 | **-0.00000** | 0.00005 | 0.00048 | 0.00074 | 0.00000 |
| | | RMSE | 0.05939 | 0.04540 | 0.03965 | 0.03511 | 0.03849 | **0.02776** |
| | | $r$ | 0.47549 | 0.67573 | 0.76395 | 0.80004 | 0.77528 | **0.87806** |
| | | SSIM | 0.76829 | 0.78370 | 0.83848 | 0.86080 | 0.84424 | **0.90728** |
| | Aug. 22, 2022 | AD | 0.00022 | **0.00000** | 0.00010 | 0.00059 | 0.00041 | 0.00001 |
| | | RMSE | 0.05309 | 0.04153 | 0.03512 | 0.03085 | 0.03459 | **0.02469** |
| | | $r$ | 0.41545 | 0.64233 | 0.76668 | 0.78018 | 0.75714 | **0.85446** |
| | | SSIM | 0.84009 | 0.84663 | 0.89628 | 0.91034 | 0.88500 | **0.93400** |
| | Oct. 01, 2022 | AD | 0.00054 | **0.00001** | 0.00014 | 0.00114 | 0.00131 | -0.00002 |
| | | RMSE | 0.07480 | 0.04748 | 0.04607 | 0.03813 | 0.04220 | **0.03335** |
| | | $r$ | 0.43262 | 0.80619 | 0.82345 | 0.86802 | 0.84504 | **0.89475** |
| | | SSIM | 0.75474 | 0.83779 | 0.85527 | 0.88651 | 0.85319 | **0.89880** |
| | Mean | RMSE | 0.05949 | 0.04365 | 0.03906 | 0.03427 | 0.03720 | **0.02817** |
| | | $r$ | 0.44408 | 0.69077 | 0.77363 | 0.80109 | 0.78476 | **0.86788** |
| | | SSIM | 0.79596 | 0.82583 | 0.86500 | 0.88490 | 0.86291 | **0.91267** |
| CIA | Aug. 03, 2022 | AD | -0.00032 | 0.00022 | -0.00000 | 0.00016 | 0.00005 | **-0.00000** |
| | | RMSE | 0.03683 | 0.03084 | 0.02296 | 0.02307 | 0.02556 | **0.01880** |
| | | $r$ | 0.44326 | 0.57986 | 0.80081 | 0.77081 | 0.70012 | **0.85148** |
| | | SSIM | 0.88829 | 0.88679 | 0.93084 | 0.92909 | 0.90728 | **0.94616** |
| | Oct. 09, 2022 | AD | -0.00010 | 0.00005 | -0.00004 | -0.00003 | 0.00008 | **-0.00002** |
| | | RMSE | 0.05379 | 0.03682 | 0.03409 | 0.03198 | 0.03264 | **0.02831** |
| | | $r$ | 0.37284 | 0.69150 | 0.77261 | 0.76910 | 0.76609 | **0.81626** |
| | | SSIM | 0.80292 | 0.85554 | 0.87727 | 0.88749 | 0.87365 | **0.90179** |
| | Dec. 01, 2022 | AD | -0.00006 | 0.00012 | -0.00003 | -0.00006 | 0.00009 | **-0.00000** |
| | | RMSE | 0.03643 | 0.02786 | 0.02332 | 0.02266 | 0.02325 | **0.02030** |
| | | $r$ | 0.38010 | 0.71589 | 0.79128 | 0.80251 | 0.79273 | **0.84024** |
| | | SSIM | 0.89898 | 0.91954 | 0.93449 | 0.93932 | 0.92816 | **0.94329** |
| | Feb. 19, 2023 | AD | -0.00025 | 0.00012 | -0.00004 | -0.00025 | 0.00011 | **-0.00000** |
| | | RMSE | 0.04433 | 0.02982 | 0.02659 | 0.02387 | 0.02594 | **0.02029** |
| | | $r$ | 0.39683 | 0.77772 | 0.82801 | 0.85720 | 0.82894 | **0.89763** |
| | | SSIM | 0.86025 | 0.91103 | 0.92091 | 0.93421 | 0.91723 | **0.94355** |
| | Mean | RMSE | 0.04285 | 0.03133 | 0.02674 | 0.02539 | 0.02685 | **0.02192** |
| | | $r$ | 0.39826 | 0.69125 | 0.79818 | 0.79991 | 0.77197 | **0.85140** |
| | | SSIM | 0.86261 | 0.89323 | 0.91588 | 0.92253 | 0.90658 | **0.93370** |

### *3.5. Effectiveness of the OL-RC and PL-RC steps*

The experiment in Section 3.3 demonstrates the effectiveness of both the OL-RC and PL-RC. In this section, we further discuss the mechanism and explain the effectiveness of these two residual compensation steps. Table 4 gives the detailed description of different types of residual maps used in this discussion. Fig. 14 and Fig. 16 show the residual maps of the NIR band and the proposed object residual index (ORI) when predicting the first image to be fused at both sites, respectively. Moreover, Fig. 15 and Fig. 17 show the enlarged sub-areas marked in the magenta rectangle in Fig. 14 (d) and Fig. 16 (d), respectively.

**Table 4** Naming and meanings of different types of residuals.

| Name | Meaning |
|---|---|
| Coarse residuals | Difference between the real coarse image at $t_p$ ($C_{tp}$) and the up-scaled OL-U prediction |
| Fine residuals | Residuals obtained by downscaling the coarse residuals using a bi-cubic interpolation (represented as $R_1$ in Section 2.3) |
| Actual residuals | Difference between the fine-scale OL-U prediction and the fine image at $t_p$ ($F_{tp}$, also reference image) |
| Object residuals | Mean actual residuals in each object |
| \|OR - FR\| | Absolute difference between the object residuals (OR) and the fine residuals (FR), a lower value represents a higher similarity between the fine residual and the object residual |
| ORI | Object residual index proposed in Section 2.3, which is used to select fine residual pixels for OL-RC |
| OL-R | Object-level residuals predicted by OL-RC, which is an approximation of the object residuals |
| OL-R + PL-R | Total residuals predicted by OL-RC and PL-RC, which is an approximation of the actual residuals |

As illustrated in Section 2.3, the coarse residuals are downscaled using a bi-cubic interpolation to get the fine residuals to improve the robustness of OL-RC in heterogeneous areas. However, as shown in sub-figure (b) of Fig. 14-Fig. 17, the fine residuals are over-smoothed by the interpolation and have poor structural information of the ground objects. The objective of OL-RC is to calculate the residual for each object by combining the fine residuals and the segmentation result, thus reconstructing a residual map that contains rich structural information of the ground objects. As shown in sub-figures (e) and (f) of Fig. 15 and Fig. 17, the high-value areas of ORI correspond to the low-value areas of |OR - FR| (see Table 4 for detailed description of |OR - FR|). This indicates that the proposed ORI can accurately

describe the similarity between the fine pixel residual and the actual object residual. With the guidance of ORI, the OL-RC selects the reliable fine residual pixels and combines the selected residuals to calculate and compensate the residual for each object. The residual map produced by OL-RC (object-level residuals, OL-R) is actually an approximation of the object residuals. As shown in sub-figure (g) of Fig. 14-Fig. 17, the OL-R predicted by OL-RC are very similar to the object residuals and contain rich structural information of the ground objects. Moreover, as shown in Table 5, the OL-R has the highest similarity ($r = 0.94165$ at the BC site, 0.96609 at the CIA site) to the object residuals compared to other types of residuals.

OL-RC assumes all fine pixels within an object have the same residual, which neglects the pixel-level spectral details of the ground objects. Moreover, the within-object land-cover changes will also reduce the accuracy of OL-RC. Therefore, the PL-RC is adopted to improve the estimation of residuals by introducing neighborhood information for each target fine pixel. By adding the PL-RC predicted residuals (PL-R) to OL-R, the predicted total residual map (OL-R + PL-R) is actually an approximation of the actual residuals. As shown in sub-figure (h) of Fig. 14-Fig. 17, the OL-R + PL-R contain more pixel-level spectral information than the OL-R, and is more similar to the actual residuals. Moreover, as shown in Table 5, the OL-R + PL-R has the highest similarity to the actual residuals compared to other types of residuals. It could also be noticed that the similarities between the OL-R + PL-R and the actual residuals are lower than those between the OL-R and the object residuals (0.82306 versus 0.94165 at the BC site, 0.87050 versus 0.96609 at the CIA site). This is because the pixel-level residuals are more complex than the object-level residuals and have a finer spatial scale, therefore, it is more difficult to predict the pixel-level residuals from the over-smoothed fine residual map.

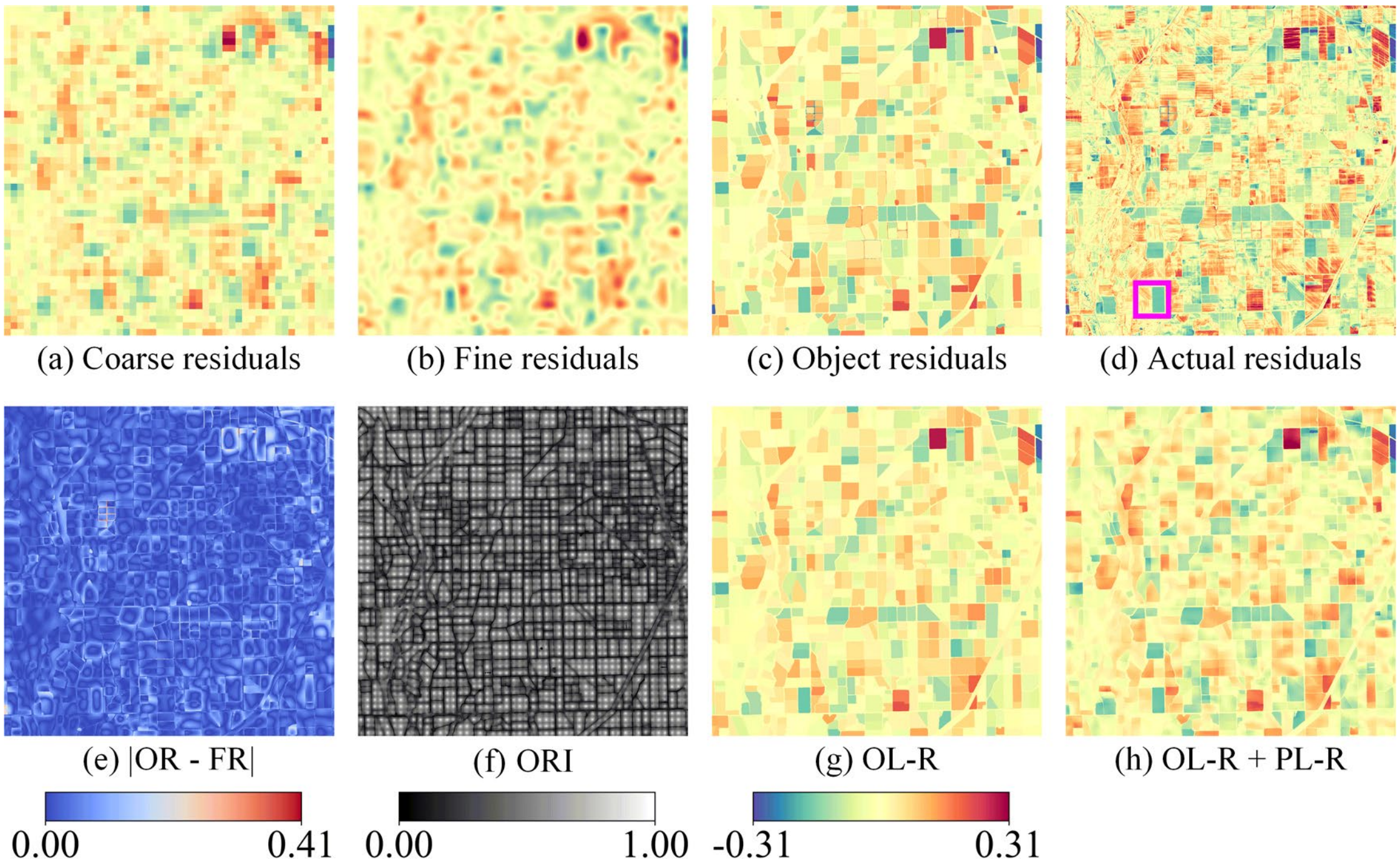

**Fig. 14.** Different types of residuals of the NIR band and the object residual index at the BC site when predicting the Sentinel-2 image on June 23, 2022. (a) Coarse residuals, (b) fine residuals ($R_1$), (c) object residuals, (d) actual residuals, (e) absolute difference between the object residuals and the fine residuals, (f) object residual index, (g) OL-R, and (h) OL-R + PL-R.

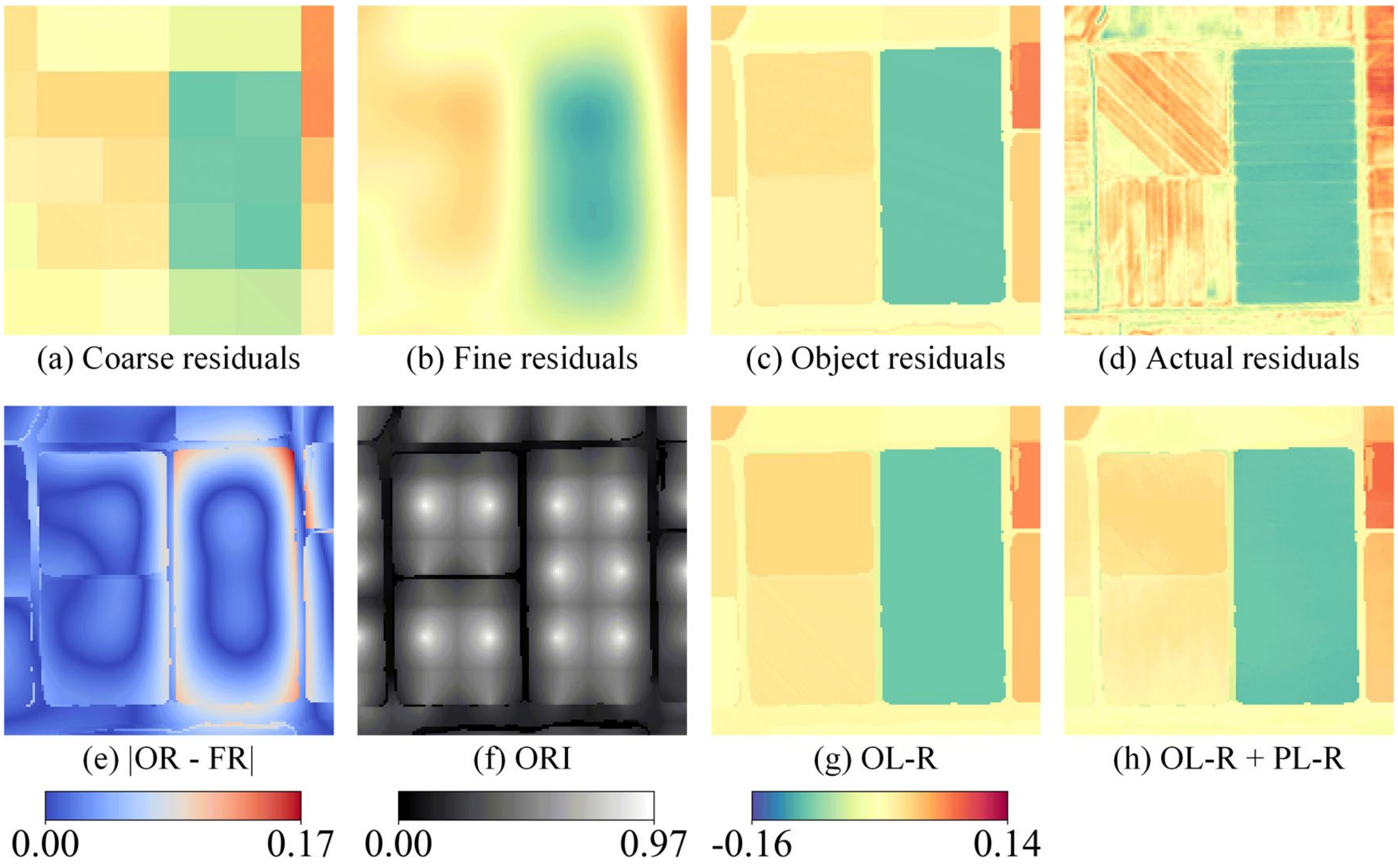

**Fig. 15.** Zoomed-in results of the sub-area marked in the magenta rectangle in Fig. 14 (d).

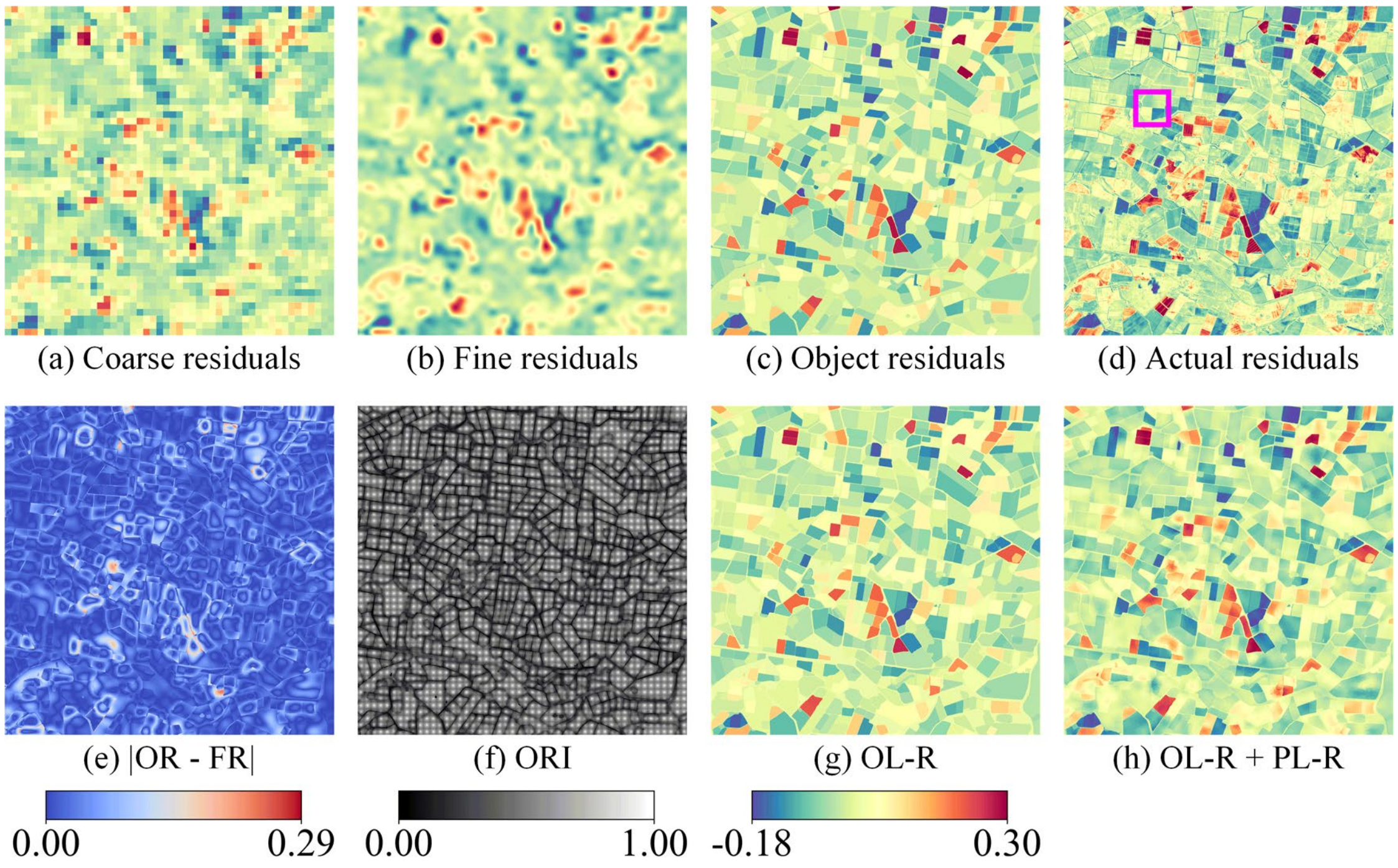

**Fig. 16.** Different types of residuals of the NIR band and the object residual index at the CIA site when predicting the Sentinel-2 image on June 23, 2022. (a) Coarse residuals, (b) fine residuals ($R_1$), (c) object residuals, (d) actual residuals, (e) absolute difference between the object residuals and the fine residuals, (f) object residual index, (g) OL-R, and (h) OL-R + PL-R.

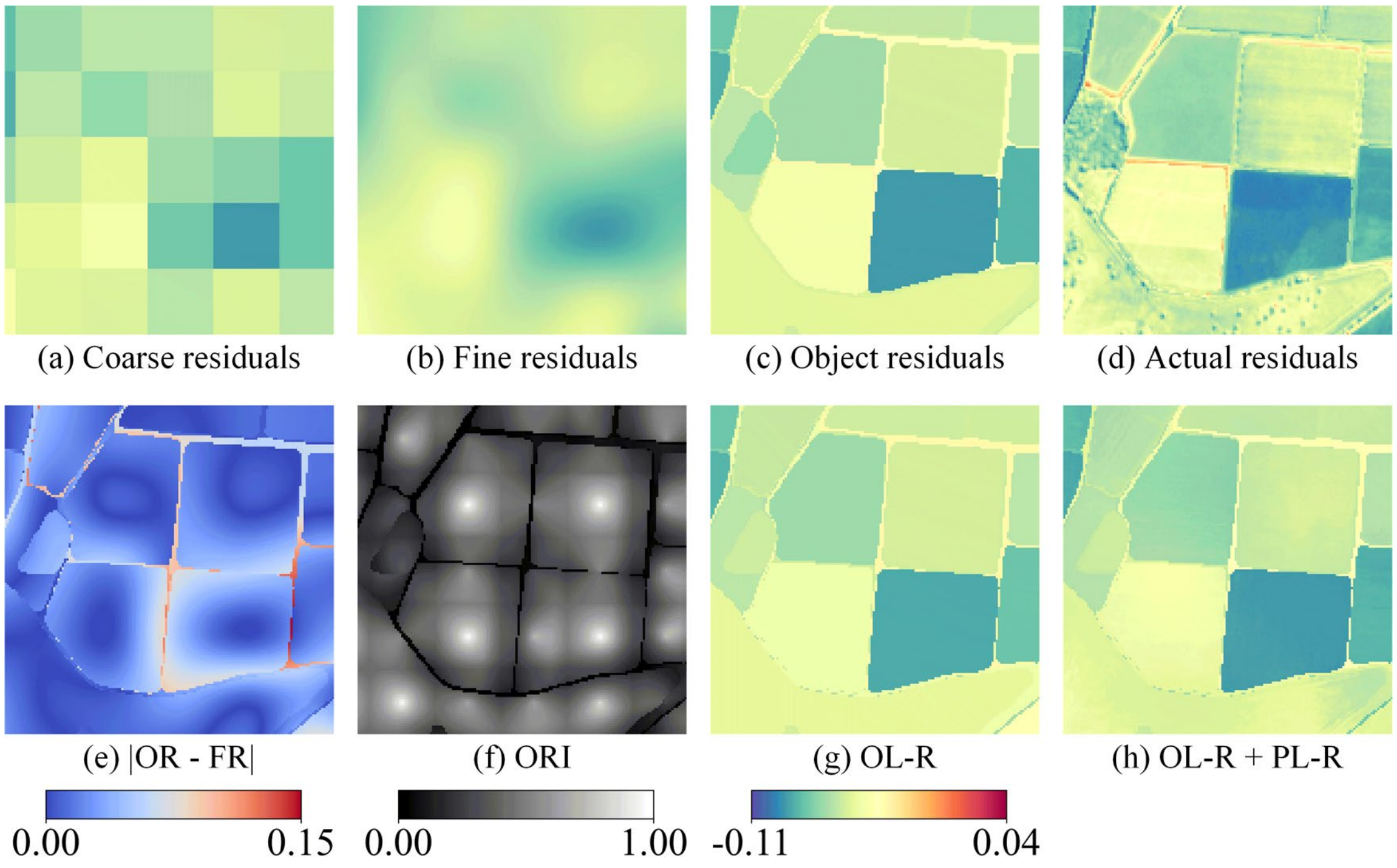

**Fig. 17.** Zoomed-in results of the sub-area marked in the magenta rectangle in Fig. 16 (d).

**Table 5** Correlation coefficient ($r$) between different residual maps and the two reference residual maps. The bold values indicate the highest accuracy in each term

| Site | Reference | Coarse residuals | Fine residuals | OL-R | OL-R + PL-R |
|---|---|---|---|---|---|
| BC | Object residuals | 0.74432 | 0.79241 | **0.94165** | 0.91381 |
| | Actual residuals | 0.73136 | 0.77685 | 0.75744 | **0.82306** |
| CIA | Object residuals | 0.76723 | 0.82533 | **0.96609** | 0.94034 |
| | Actual residuals | 0.77410 | 0.83138 | 0.81535 | **0.87050** |

## 4. Examples of applications supported by OBSUM

As illustrated in Section 3.4 and Section 3.5, OBSUM can retrieve strong temporal changes and recover land-cover changes accurately. Therefore, OBSUM has great potential to support various remote sensing applications. Crop progress monitoring (Gao et al., 2017; Gao et al., 2015) and land-cover classification (Li et al., 2022; Zhang et al., 2018) are two important application scenarios of spatiotemporal fusion, and the capabilities of OBSUM in these two applications are discussed below.

### *4.1. Crop progress monitoring*

Monitoring the crop progress through the growing season is of great significance for precision agricultural management and understanding vegetation responses to climate change (Zhang et al., 2003). Spatiotemporal fusion provides a solution to improve both the spatial and temporal resolution of crop progress monitoring. We collected 349 valid MOD09GA surface reflectance records and 17 cloud-free Landsat 9 Collection 2 Level-2 surface reflectance observations of the BC site in 2022. The MOD09GA images were first reprojected using the MODIS Conversion Toolkit (MCTK), and then resampled to the spatial resolution of 480 m. Moreover, the 480 m cloud masks of MOD09GA images were also extracted and then used to mask the observations contaminated by clouds in the following fusion process.

For each MOD09GA image, the Landsat 9 image that has the closest time interval to it was selected as the base image for predicting the 30 m Landsat-like image using OBSUM. After that, the fused image was used to calculate the Enhanced Vegetation Index (EVI) for monitoring the progress of rice. Compared to other vegetation indices, EVI is more sensitive to the high biomass and varies greatly throughout the growth period, which could benefit the monitoring of crop progress (Zhao et al., 2023). Considering the gaps in the OBSUM predictions and the EVI maps that were caused by cloud contamination in the MOD09GA observations, we applied the vegetation index Savitzky-Golay filter (Chen et al., 2004) to reconstruct a high-quality EVI time-series. Fig. 18 illustrates the spatiotemporal fusion of the Landsat image and the cloudy MOD09GA image using OBSUM and the reconstruction of the gap-free EVI map at date of year (DOY) 193, 2022.

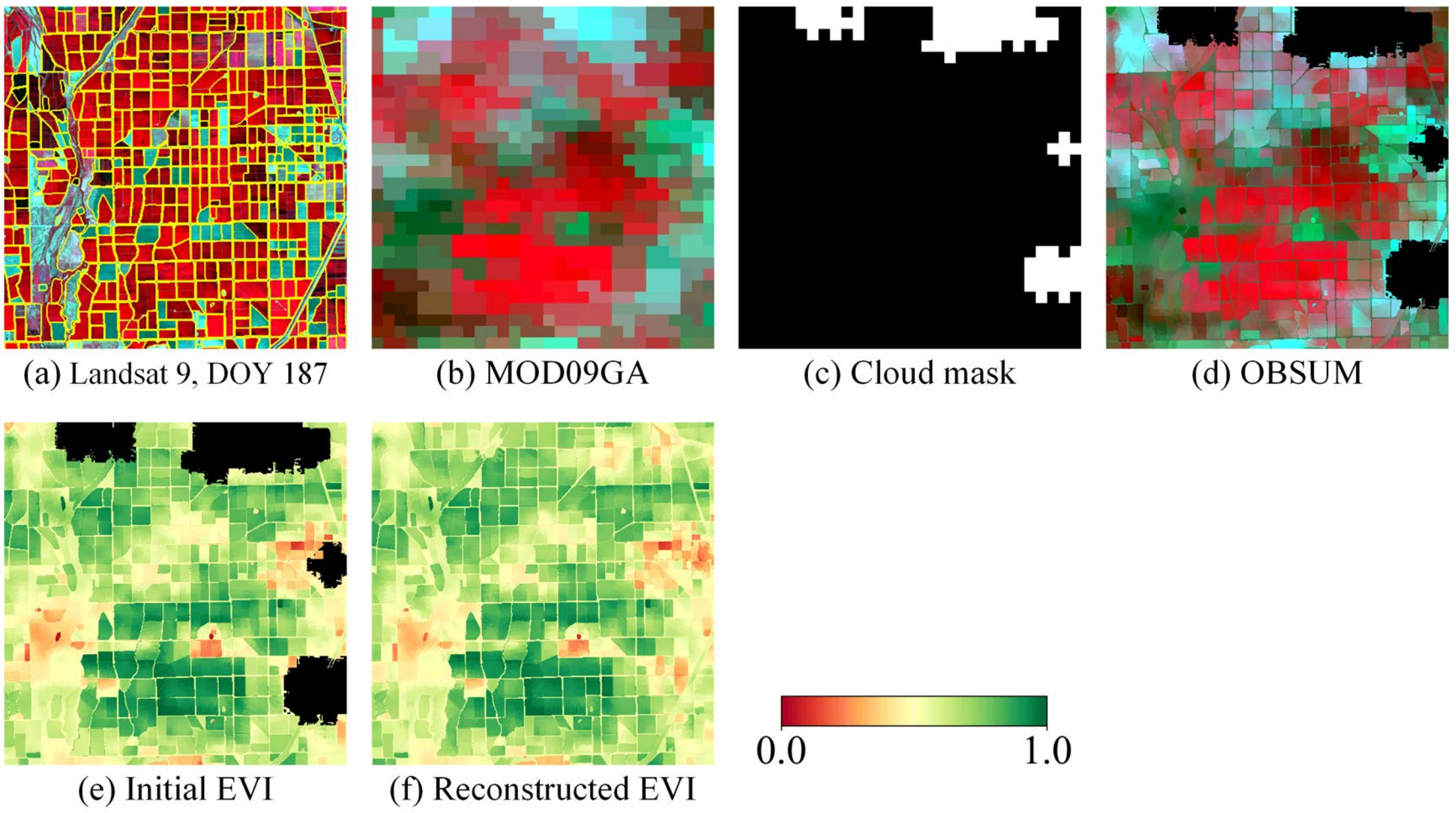

**Fig. 18.** Input images, intermediate images, and reconstructed gap-free EVI map at DOY 193, 2022. (a) Landsat 9 image at Doy 187 and the segmentation result, (b) MOD09GA image at DOY 193, (c) MOD09GA cloud mask, (d) Landsat-like image predicted by OBSUM, (e) EVI map with gaps, and (f) reconstructed gap-free EVI map.

The USDA's 30 m Cropland Data Layer (CDL) (USDA, 2022b) covering the BC site in 2022 was collected for extracting the rice pixels in the fused Landsat-like images. After that, the mean EVI value of the rice pixels was calculated for each DOY. The EVI values were then analyzed by the TIMESAT (Jonsson & Eklundh, 2002) software to extract the seasonality parameters, including the start of season, middle of season, end of season, and length of season. Fig. 19 shows the EVI curve smoothed by the double logistic function and the extracted seasonality parameters. According to the TIMESAT analysis, the growing season of rice at the BC site in 2022 started on May 25 (DOY 145), reached its peak at July 26 (DOY 207), ended at Oct. 26 (DOY 299), and the season length was 154 days. We also collected the Crop Progress (CP) report (USDA, 2022c) released by the USDA as the reference for crop progress. The state-level CP report is publicly available at a weekly frequency. According to the CP report, by May 29 (DOY 149), 95% of the rice in California had been planted, and 50% of the rice had been emerged. By July 24 (DOY 205), 40% of the rice in California had been headed, with 75% in good condition and 20% in excellent condition. By Oct. 23 (DOY 296), 75% of the rice in California had been harvested, and 90% by Oct. 30 (DOY 303). The seasonality parameters obtained by OBSUM plus TIMESAT are generally consistent with those of the CP report, which indicates the potentiality of OBSUM for supporting crop progress monitoring at a fine scale.

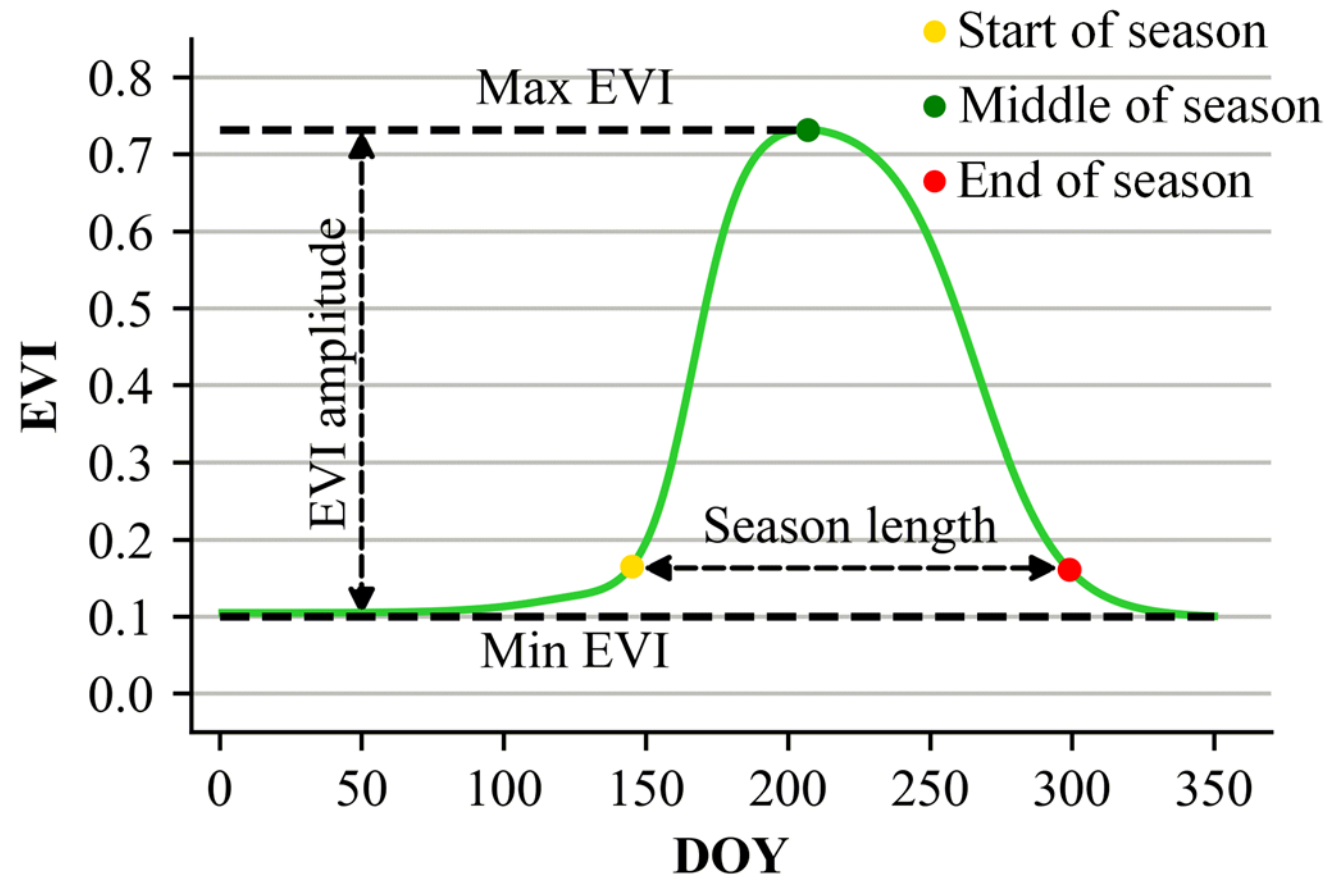

**Fig. 19.** Smoothed EVI curve and the seasonality parameters extracted using TIMESAT.

Fig. 20 shows the reconstructed gap-free EVI time-series at the BC site from the start of season (COY 145) to the end of season (DOY 299) in 2022, and the time intervals between the EVI maps is two weeks. The phenology of rice and the Landsat-like spatial details can be clearly observed in the EVI maps. The above discussion suggests that OBSUM can support dynamic monitoring of crop progress at a fine field-scale.

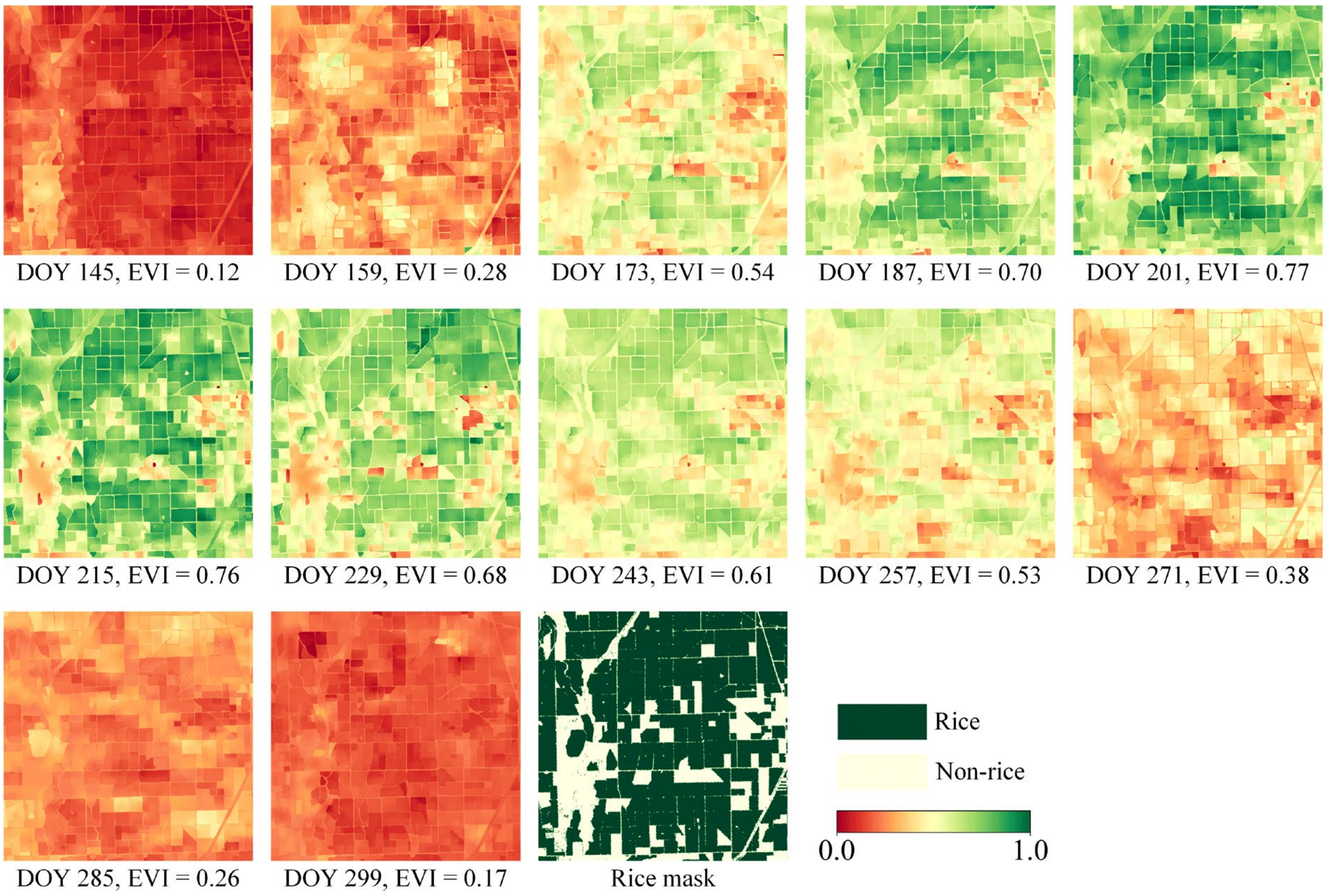

**Fig. 20.** Reconstructed gap-free EVI time-series at the BC site from DOY 145 to DOY 299 in 2022. The rice mask is extracted from the Crop Data Layer provided by the USDA.

### *4.2. Land-cover classification*

Land-cover classification is another important application of spatiotemporal fusion. Considering the land-cover dynamics, the real image and the fused images at the CIA site on Aug. 03, 2022 were used to perform the classification experiment. The real Sentinel-2 image was used to manually select the training and validation labels for building a simple support vector machine (SVM) classifier and evaluating the classification maps, respectively. Moreover, the real Sentinel-2 image was also classified into five land-cover classes using

SVM to get a reference classification map for visual comparison. Fig. 21 shows the real Sentinel-2 image, the selected training and validation labels, the SVM classification results of the images fused by different spatiotemporal fusion methods, and the reference classification map. Moreover, the enlarged sub-areas marked in the yellow rectangle in Fig. 21 (a) are shown in Fig. 22. One can observe in Fig. 22 that UBDF, STARFM, Fit-FC, and OBSTFM failed to accurately predict vegetation 2 in the classification maps. Moreover, STARFM and FSDAF failed to distinguish the complete extent of the water in the classification maps. The classification result of the image fused by OBSUM is the most similar to the reference map in terms of both class labels and extent of ground objects. The reason is that OBSUM contains two object-level processes (OL-U and OL-RC) that can preserve the structural information of the ground objects, as well as obtain high fusion accuracy.

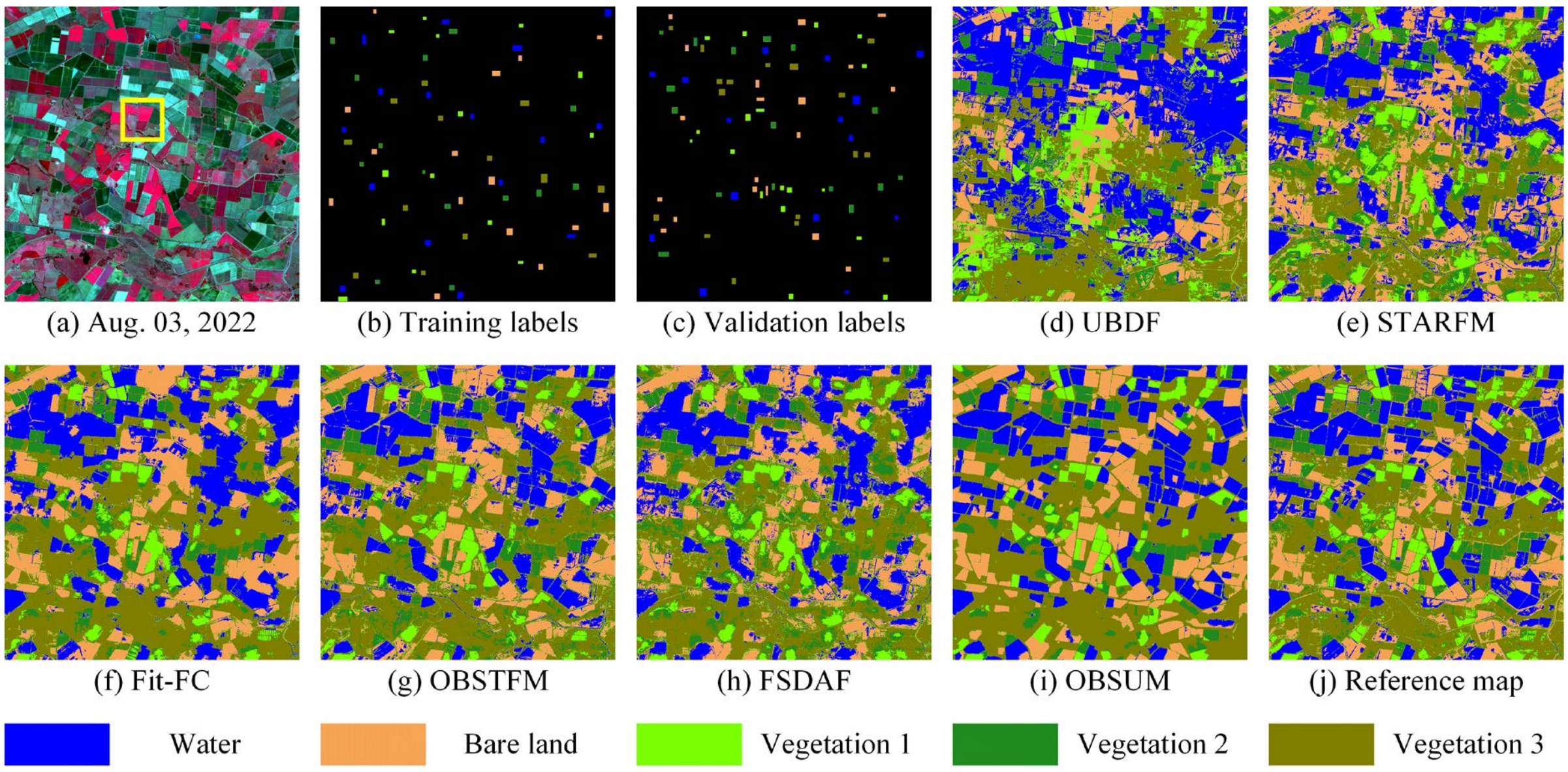

**Fig. 21.** Classification results at the CIA site on Aug. 03, 2022. (a) Real Sentinel-2 image, (b) training labels, (c) validation labels, (d)-(i) SVM classification maps of the images fused by UBDF, STARFM, Fit-FC, OBSTFM, FSDAF, and OBSUM, and (j) reference classification map.

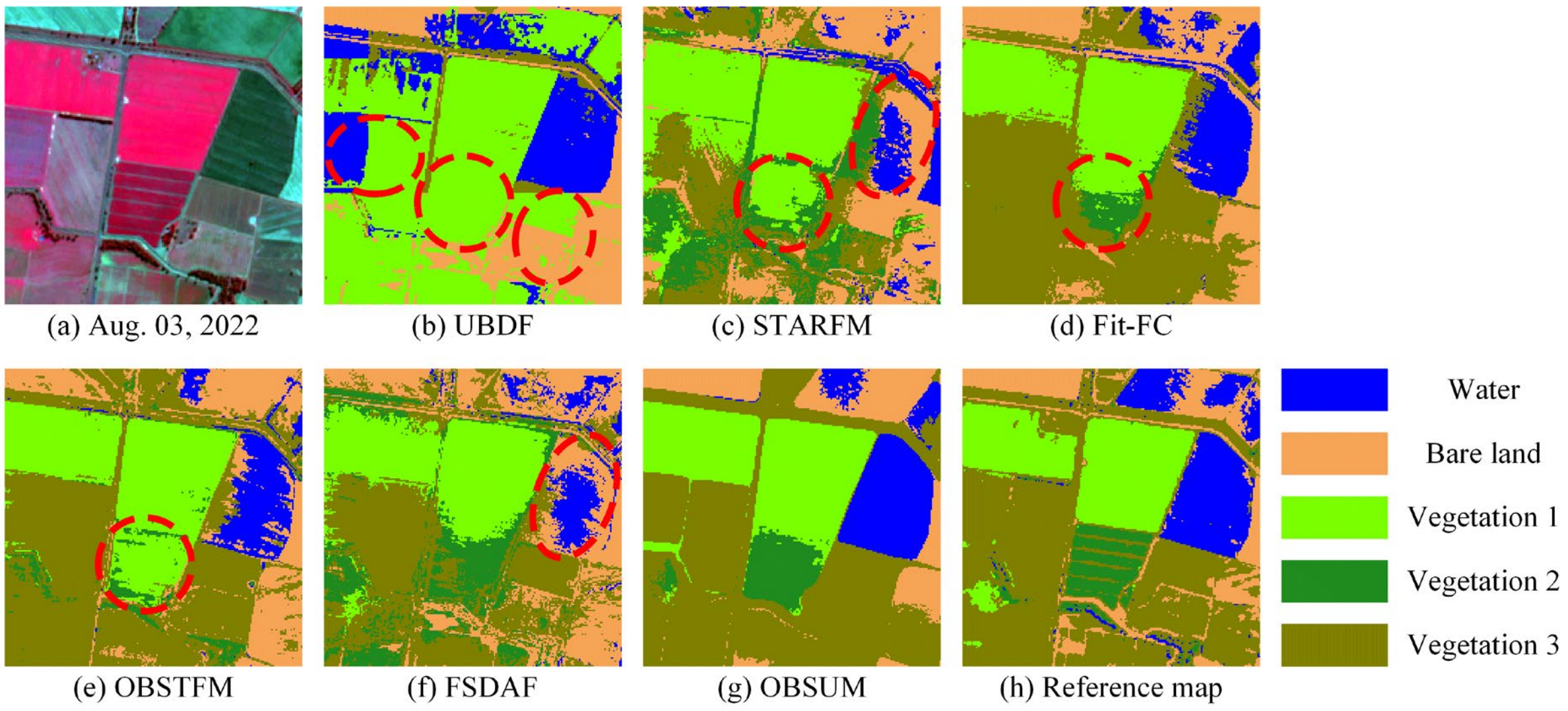

**Fig. 22.** The corresponding zoomed-in results of the sub-area marked in the yellow rectangle in Fig. 21 (a). The ellipses represented by the red dashed line show the mis-classification in the fused images.

Table 6 shows the overall accuracy (OA) and Kappa coefficient (Kappa) of the classification maps obtained by different spatiotemporal fusion methods in the land-cover classification task. The accuracies were calculated using the validation labels shown in Fig. 21 (c). It can be observed that OBSUM achieved the highest OA and Kappa coefficient among all methods, and the two indices of OBSUM are similar to those of the reference map (the classification result directly obtained from the real Sentinel-2 image), which indicates the superior performance of OBSUM in the application of land-cover classification.

**Table 6** Accuracy of different spatiotemporal fusion methods in the land-cover classification task. The bold values indicate the highest accuracy in each term.

| Metric | Reference map | Method | | | | | |
|---|---|---|---|---|---|---|---|
| | | UBDF | STARFM | Fit-FC | OBSTFM | FSDAF | OBSUM |
| OA | 0.9711 | 0.4896 | 0.7659 | 0.9438 | 0.8618 | 0.8822 | **0.9716** |
| Kappa | 0.9637 | 0.3565 | 0.7058 | 0.9294 | 0.8264 | 0.8520 | **0.9644** |

## 5. Discussion and conclusion

This study proposed an object-based spatial unmixing model (OBSUM) for spatiotemporal fusion of remote sensing images. OBSUM predicts the fine image at the prediction date by blending one fine image at the base date and one coarse image at the prediction date. It includes three indispensable fusion steps: object-level unmixing (OL-U), object-level residual compensation (OL-RC), and pixel-level residual compensation (PL-RC). The OL-U produces an initial fusion result by incorporating spatial unmixing and object-based image analysis. After that, the OL-RC calculates and compensates the residual for each object, which can significantly recover the spectral information. Finally, the PL-RC is applied to retrieve the within-object land-cover changes, thus further improving the fusion accuracy.

In the preprocessing step, the fine image at the base date is classified into several land-cover classes to define the endmembers, and also calculate the endmembers' fractions for spatial unmixing. The fine image is also segmented by SAM to define the ground objects, thus guiding the subsequent object-level fusion steps. More importantly, the classification result and the segmented image objects are used together to get the object-level land-cover classification map. This can eliminate the pixel-level classification errors introduced by intra-class spectral variation and provide a classification map that can be applied to the OL-U. In OL-U, the fine image, coarse image, and image objects are utilized together to obtain an initial fusion result in which no block effect exists. In the following OL-RC, the coarse image at the prediction date is used to calculate the residuals, then downscaled to fine-scale. The object residual index (ORI) map generated from the image objects is applied to guide the residual compensation. In the final PL-RC, the coarse image is used again to calculate the residuals, and the fine image is used to select similar pixels to strengthen the prediction.

Benefiting from the utilization of all available input images and object-level information, one noticeable advantage of OBSUM is that it needs the minimum number of input images, i.e., one fine image at the base date and one coarse image at the prediction date. In the implementation of spatiotemporal fusion methods, one difficulty is that sometimes there is no matching fine-coarse image pair at the base date due to cloud contamination or a long revisit time (Goyena et al., 2023). OBSUM can be used without the need for one coarse image at the base date, and this is of great significance in promoting the application of spatiotemporal fusion under complex conditions and in improving the implementation flexibility. Moreover, OBSUM can also be used when the fine and coarse images have inconsistent spectral bands. Therefore, OBSUM has great potential for downscaling the total 21 spectral bands of Sentinel-3 OLCI, which range from 400 nm to 1020 nm. The downscaled Sentinel-2-like images provide more detailed spectral information and can facilitate a large number of downstream applications (Tang et al., 2021).

The proposed OBSUM was tested at two typical agricultural sites and compared with five methods, including UBDF, STARFM, Fit-FC, OBSTFM, and FSDAF. The experimental results demonstrate that OBSUM can retrieve both phenological changes and land-cover changes, thus outperforming the comparison methods in terms of visual effect and four accuracy indices. Benefiting from its accurate fusion performance, OBSUM has great potential for supporting various remote sensing applications at a fine spatial scale, such as dynamic crop progress monitoring and land-cover classification.

It is noteworthy that there are also some limitations that motivate our future improvement on OBSUM. First, OBSUM directly unmixes the coarse image at the prediction date $t_p$ in the OL-U process instead of the temporal change between the coarse images ($\Delta C = C_{t_p} - C_{t_b}$). Moreover, despite the classification map refinement and the OL-U can avoid block effects in the fused images, they may also ignore the small objects that have not been properly segmented. As a result, the OBSUM prediction has relatively poor pixel-level textural features compared to the results of STARFM, Fit-FC, OBSTFM, and FSDAF, which distribute the temporal change to the fine image (see the visual comparison in Fig. 11 and Fig. 12). In our further research, the coarse image at the base date $t_b$ will be integrated into OBSUM to unmix the temporal change (similar to the unmixing strategy in STDFA), thus improving the local texture information in the fused image. Besides, OBSUM will also be extended to take multiple image pairs as input, thus further improving the fusion accuracy.

Second, among the three fusion steps of OBSUM, only the PL-RC is designed for recovering within-object land-cover changes. Therefore, OBSUM may fail to capture abrupt land-cover changes that would break the object boundaries segmented from the fine image at $t_b$, such as floods, forest degradation, wildfires, etc. The main difficulties in such fusion circumstances are to select the unchanged coarse pixels for spatial unmixing and to identify the precise extent of land-cover changes (Zhu et al., 2018). In our further research, the reliability index (Shi et al., 2022), land-cover change detection (Jiang & Huang, 2022), and thin plate spline (TPS) interpolation (Zhu et al., 2016) techniques will be integrated into the OBSUM framework. The reliability index can filter out the changed and unreliable coarse pixels, thus improving the local unmixing step's robustness to land-cover changes. Moreover, it can also guide the residual compensation to obtain a more accurate prediction. Change detection and TPS interpolation can help to retrieve the abrupt land-cover changes and predict the changed object boundaries, respectively.

The Python code of OBSUM and the experimental dataset are available at https://github.com/HoucaiGuo/OBSUM-code.

## Declaration of Competing Interest

The authors declare that they have no known competing financial interests or personal relationships that could have appeared to influence the work reported in this paper.

## Acknowledgement

This study was supported by the China Scholarship Council under Grant 202306860011. The authors would like to thank Dr. Feng Gao, Prof. Qunming Wang, Mr. Dizhou Guo, Prof.